%% file: acl_latex.tex
\pdfoutput=1

\documentclass[11pt]{article}

\usepackage[final]{acl}

\usepackage{times}
\usepackage{latexsym}
\usepackage{amsmath}
\usepackage{hyperref}
\usepackage{booktabs}
\usepackage{adjustbox}
\usepackage{comment}
\usepackage{makecell}
\usepackage{multirow}
\usepackage{colortbl}
\usepackage{float}
\usepackage{placeins}
\usepackage{subcaption}
\usepackage{algpseudocode}
\usepackage{algorithm}

\usepackage{todonotes}
\usepackage{url}

\usepackage[capitalize, nameinlink]{cleveref}
\crefformat{section}{Section #2#1#3} %
\crefformat{subsection}{Section #2#1#3}
\crefformat{subsubsection}{Section #2#1#3}
\crefformat{figure}{Figure #2#1#3}
\crefformat{appendix}{App. #2#1#3}

\definecolor{ForestGreen}{RGB}{34,139,34}
\definecolor{electricindigo}{rgb}{0.44, 0.0, 1.0}
\definecolor{cobalt}{rgb}{0.8, 0.28, 0.8}
\definecolor{BurntOrange}{RGB}{204, 85, 0}
\definecolor{RoyalBlue}{RGB}{65, 105, 225}

\usepackage[T1]{fontenc}

\usepackage[utf8]{inputenc}

\usepackage{microtype}
\usepackage{svg}

\usepackage{inconsolata}

\usepackage{graphicx}

\usepackage{pythonhighlight}

\title{Stress-testing Machine Generated Text Detection: \\ Shifting Language Models Writing Style to Fool Detectors}

\author{
  {\bf Andrea Pedrotti\textsuperscript{$\alpha$}, Michele Papucci\textsuperscript{$\beta$,$\gamma$}, Cristiano Ciaccio\textsuperscript{$\gamma$},} \\ {\bf Alessio Miaschi\textsuperscript{$\gamma$}, Giovanni Puccetti\textsuperscript{$\alpha$}, Felice Dell'Orletta\textsuperscript{$\gamma$}, Andrea Esuli\textsuperscript{$\alpha$}}\\
  \textsuperscript{$\alpha$} Istituto di Scienza e Tecnologie dell'Informazione ``A. Faedo'' (CNR-ISTI) \\ 
  \texttt{\{name.surname\}@isti.cnr.it} \\
  \textsuperscript{$\beta$} Department of Computer Science, University of Pisa \\
  \textsuperscript{$\gamma$} ItaliaNLP Lab, Istituto di Linguistica Computazionale ``Antonio Zampolli'' (CNR-ILC) \\ \texttt{\{name.surname\}@ilc.cnr.it}
}

\begin{document}
\maketitle

\begin{abstract}
Recent advancements in Generative AI and Large Language Models (LLMs) have enabled the creation of highly realistic synthetic content, raising concerns about the potential for malicious use, such as misinformation and manipulation. Moreover, detecting Machine-Generated Text (MGT) remains challenging due to the lack of robust benchmarks that assess generalization to real-world scenarios. In this work, we present a pipeline to test the resilience of state-of-the-art MGT detectors (e.g., Mage, Radar, LLM-DetectAIve) to linguistically informed adversarial attacks. To challenge the detectors, we fine-tune language models using Direct Preference Optimization (DPO) to shift the MGT style toward human-written text (HWT). This exploits the detectors' reliance on stylistic clues, making new generations more challenging to detect. Additionally, we analyze the linguistic shifts induced by the alignment and which features are used by detectors to detect MGT texts. 
Our results show that detectors can be easily fooled with relatively few examples, resulting in a significant drop in detection performance. This highlights the importance of improving detection methods and making them robust to unseen in-domain texts. We release code, models, and data to support future research on more robust MGT detection benchmarks\footnote{\url{https://github.com/gpucce/control_mgt}.}.

\end{abstract}

\section{Introduction}
\label{sec:intro}

Recent advancements in Generative AI and Large Language Models (LLMs) have led to the development of systems, such as GPT-4 \citep{OpenAI_2023_GPT-4_Technical_Report}, Claude\footnote{\href{https://www-cdn.anthropic.com/de8ba9b01c9ab7cbabf5c33b80b7bbc618857627/Model_Card_Claude_3.pdf}{anthropic.com/claude-model-card}}, Llama 3 \citep{dubey2024llama} and DeepSeek V3 \citep{deepseekai2024deepseekv3technicalreport} among others, that can generate text that is often indistinguishable from human-written content \citep{Dugan_Ippolito_Kirubarajan_Shi_Callison-Burch_2023}.

\begin{figure}[t!]
    \centering
    \includegraphics[width=1\linewidth]{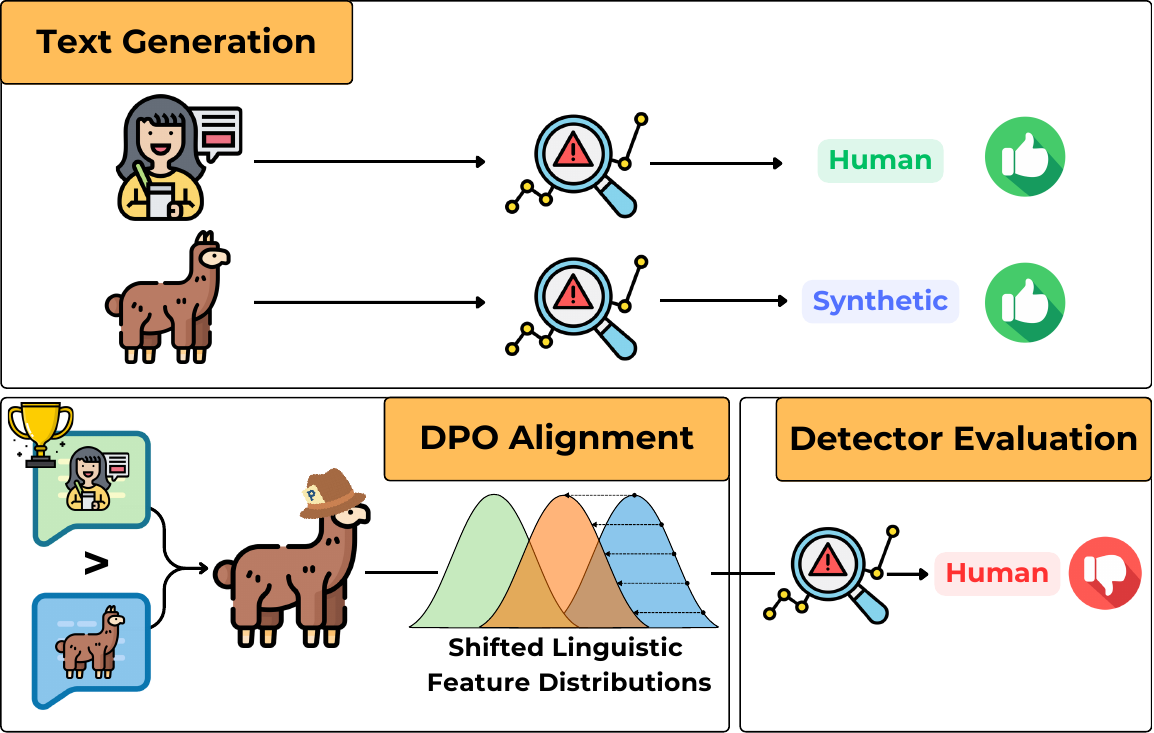}
    \caption{Overview of the proposed methodology. We fine-tune LLMs to generate MGTs that align stylistically with human-written texts (HWTs) by shifting their linguistic feature distributions via DPO (\textit{DPO Alignment}). Then, we evaluate the effectiveness of this alignment against MGT detectors (\textit{Detector Evaluation}).}
    \label{fig:graph-abstract}
\end{figure}

This capability, along with the many beneficial applications of LLMs, also enables malicious actors to generate synthetic content for deceptive purposes. For example, it can be used to manipulate online traffic and spread misinformation through content farms \citep{Puccetti_2024} or to influence human revisions of sensitive documents in critical domains, such as scientific peer review\footnote{\href{https://www.aclweb.org/adminwiki/index.php/ACL_Policy_on_Publication_Ethics}{aclweb.org/genai-peerreview-guidelines}}. A notable case is Galactica \citep{taylor2022galacticalargelanguagemodel}, a language model designed for scientific writing assistance, which faced strong criticism from the scientific community over concerns about potential misuse, ultimately leading to its withdrawal\footnote{\href{https://www.technologyreview.com/2022/11/18/1063487/meta-large-language-model-ai-only-survived-three-days-gpt-3-science/}{technologyreview.com/galactica-shut-down}}.

To mitigate concerns about the undisclosed use of machine-generated text (MGT), the development of reliable detection methods is essential for a responsible deployment of generative AI, and many detectors have been proposed. However, ensuring the trustworthiness of MGT detectors requires robust benchmarks to evaluate their performance. While several efforts have been made to create comprehensive and long-lasting benchmarks \citep{wang-etal-2024-m4}, many of them tend to saturate quickly: for instance, in shared tasks on MGT detection \cite{wang-etal-2024-semeval-2024, dugan2025genaicontentdetectiontask}, top-ranked systems often achieve near-perfect performance, like the first-place participant in \citet{wang-etal-2024-semeval-2024} that reported an overall accuracy above 96\%. Despite these results, \citet{doughman-etal-2025-exploring} highlight that most detectors struggle when applied to out-of-domain (OOD) samples, attributing this weakness to the detectors' reliance on superficial stylistic cues and their sensitivity to variations in text complexity, leading to high accuracy in controlled settings but poor generalization when identifying MGT in diverse, real-world texts.

In this work, we take a first step toward addressing the linguistic shortcut learning exhibited by MGT detectors. In particular, we propose a pipeline that identifies the linguistic properties leveraged by current MGT detectors to perform their classification, and use this insight to fine-tune existing LLMs with Direct Preference Optimization (DPO) \citep{dpo2023}, enabling them to generate synthetic texts that closely resemble human-written texts (HWT) from a writing style point-of-view. The motivation behind this approach is that, as observed in previous studies \citep{krishna-etal-2020-reformulating,esuli2024you}, the distributions of linguistic phenomena between MGT and HWT are different.

We hypothesize that if MGT detectors rely on ``linguistic shortcuts'', created by this difference in distributions, we can fool them by aligning the writing style of LLMs more closely to the human writing style, producing texts stylistically akin to HWT by aligning the linguistic profile of MGTs to HWTs (see \Cref{fig:graph-abstract}). 
These newly generated texts can be used as a more robust, trustworthy, and less domain-dependent benchmark for MGT detectors, as they closely resemble the HWT style, making it difficult for detectors to rely on superficial stylistic cues.

Our contributions are summarized below:
\begin{enumerate}
    \item We \textbf{develop a pipeline} for generating synthetic texts that are harder to detect by aligning LLMs to HWT via DPO. Furthermore, our pipeline can be easily adapted to work with other LLMs, eventually fostering the development of more robust, generalizable detection models.\\
    \item We \textbf{evaluate a suite of state-of-the-art MGT detectors} and human raters when challenged with our newly generated in-domain adversarial examples. Within this evaluation process, we highlight how current MGT detectors rely on shallow linguistic cues, resulting in a drop in performance. %
    \item We \textbf{present and in-depth analysis of the linguistic feature exploited} by detectors and human raters when identifying MGTs and find only a limited intersection between the two sets;
    \item By applying our methodology, we \textbf{develop} a set of challenging adversarial datasets for testing the robustness of current state-of-the-art MGT detectors\footnote{Generator models will be available for research purposes upon request.}.
\end{enumerate}

\section{Methodology}
\label{approach}

To make MGTs more challenging, we propose an adversarial methodology that leverages human and synthetic texts to align a model to generate texts with a writing style that's more similar to a human one. 

To explore this, we focus on two domains where MGT could have a significant societal impact: news articles (i.e. BBC) and scientific writing (i.e. paper abstracts). Our adversarial pipeline is described in Algorithm \ref{alg:cap}.

\begin{algorithm}[t!]
\caption{Pipeline for Adversarial Evaluation of Detectors}\label{alg:cap}
\begin{algorithmic}[1]
\State Select a Dataset $\mathcal{D}$ of HWT;
\State Select an LLM $\mathcal{M}$;
\State Sample MGT $\sim\mathcal{M}$, using titles in $\mathcal{D}$ as prompts for the generation;
\State We obtain $\mathcal{D}_{par} = $ (HWT, MGT) parallel dataset;
\State Evaluate state-of-the-art Detectors on $\mathcal{D}_{par}$;
\State Choose pairs of HWT and MGT from $\mathcal{D}_{par}$;
\State Fine-tune $\mathcal{M}$ using DPO with the selected pairs, tagging the HWT as the preferred answer; We obtain $\mathcal{M}'$, a model that generates text that aligns more closely with HWT.
\end{algorithmic}
\end{algorithm}
\noindent This can be iterated multiple times by setting $\mathcal{M} = \mathcal{M}'$ and starting over from step $3$. 
When collecting the samples for adversarial training (step $6$), we experiment with different data selection strategies to fine-tune models that are ``harder to detect'' (see Sec. \ref{sec:dpo-align}).

\subsection{DPO Training for Adversarial Alignment}
\label{sec:dpo-align}

DPO \citep{dpo2023} is a reinforcement learning technique used for model alignment, that instead of fitting a reward model adjusts model weights directly based on human preference. 
To steer the model's generation stylistically towards HWT, we leverage DPO by creating preference datasets where the preferred options are HWT and the dispreferred ones are MGT. We test two different approaches to build a preference dataset. In the first one, which we call \emph{dpo}, we take a set composed of HWT and MGT that are generated by the model we want to align. We label all the HWT as the preferred ones.

In the second one, which we call \emph{dpo-ling}, we select couples of (HWT, MGT) in a linguistically informed way: using an SVM classifier trained on explicit linguistic features extracted from the texts to identify MGT, we select the top ten most discriminating features. Then, for each of those features, we take the top-$k$ pairs where the absolute distance on that feature between the HWT and MGT is the largest. This second technique allows us to see both whether detectors use specific linguistic features to identify MGT, and how well DPO can steer the model's generation writing style in a way that follows given linguistic constraints.

Then, we use the model aligned with either dataset (\emph{dpo} and \emph{dpo-ling}) to generate a new set of MGT. These new generations can be used to create a second iteration of these datasets to align the model further. This can be iterated any number of times if you have sufficient data. However, for both techniques, we avoid selecting the same couples for more than one iteration, and for \emph{dpo-ling} we always select a different set of linguistic features. See \Cref{app:train-dpo} for more details.

\paragraph{Linguistic Features} The set of linguistic features we use for training the SVM-based detector, as well as for extracting and deriving the \textit{dpo-ling} training set, are extracted with ProfilingUD \cite{brunato-etal-2020-profiling}, a tool that allows the extraction of raw, morpho-syntactic and syntactic levels of annotation based on the UD formalism \cite{10.1162/coli_a_00402}. These features have been successfully used for text classification tasks, as well as to evaluate the capabilities of LLMs to adhere to specific linguistic constraints \cite{miaschi-etal-2024-evaluating,ciaccio2024controllable}.

\section{Experimental Setup}
\label{sec:experimental_setup}

\begin{table*}[ht]
    \centering
    \input{tables/all_results_horizontal}

    \caption{Macro Average F1-score on XSUM balanced test split (45.000 documents) and arXiv Abstracts test split (8.000 documents). The symbol dagger (\textdagger) denotes a method explicitly trained on our dataset. The \colorbox{gray!30}{grey rows} report the original models and \textbf{bold} values denote the best result across generator models.} %
    \label{tab:res-xsum}
\end{table*}

For assessing the effectiveness of our pipeline in generating ``harder to detect'' machine generated texts, we evaluate four state-of-the-art LLM-based detectors: \textit{(i)} \textbf{RADAR} \cite{hu2023radar}: a RoBERTa-large-based detector trained using an adversarial learning setup where a paraphraser rewrites machine-generated text to simulate paraphrasing attacks. The model is trained on WebText-derived human data and machine-generated completions generated by 8 language models; \textit{(ii)} \textbf{MAGE} \cite{li-etal-2024-mage}: a Longformer-based detector fine-tuned on human and machine-written texts generated by 27 LLMs across 7 diverse writing tasks, including news (1000 examples from the XSum dataset) and abstracts of scientific articles; \textit{(iii)} \textbf{LLM-DetectAIve}\footnote{This model is trained on a four-class classification problem for mixed human and machine written texts, we aggregate them into binary classification.} \citep{abassy-etal-2024-llm}: a DeBERTa-based detector trained across multiple domains (excluding news) on the M4GT-Bench dataset which is augmented with machine-generated text obtained from several LLMs, including Llama3-8b; and \textit{(iv)} \textbf{Binoculars} \citep{hans2024spotting}: a zero-shot LLM detector approach that leverages two Falcon-based LLMs to compute a normalized perplexity metric.
Additionally, we test an in-domain fine-tuned \textbf{RoBERTa} and a \textbf{Support Vector Machine} (SVM) with a linear kernel using the set of linguistic features previously described.

As discussed in \Cref{sec:sota}, a key challenge in developing MGT detectors is addressing domain shift. To account for this in our experiments, we test our approach on two datasets from sensitive domains, news and scientific writing: (i) XSUM \citep{narayan_2018_xsum}, a large dataset of around 200k news articles from the BBC. In particular, we focus on a random subset of 100k news. Notably, a portion of XSUM is included in the training set of the MAGE detector, making it a useful proxy for evaluating in-domain MGT detection. (ii) arXiv Abstracts\footnote{\href{https://www.kaggle.com/datasets/spsayakpaul/arxiv-paper-abstracts}{kaggle.com/arxiv-paper-abstracts}}, a large dataset of arXiv Abstract with the title of the paper, which is partly included in M4 \citep{wang-etal-2024-m4}, a large scale benchmark dataset for machine-generated text detection. We focus on a random sample of 20.000 abstracts.

For the generation of the synthetic texts, we leverage two instruction-based models, the 8B version of LLaMA 3.1 \cite{dubey2024llama},\footnote{\href{https://huggingface.co/meta-llama/Llama-3.1-8B-Instruct}{huggingface.co/Llama-3.1-8B-Instruct}} and the 2B version of Gemma 2 \cite{gemmateam2024gemma2improvingopen}.\footnote{\href{https://huggingface.co/google/gemma-2-2b-it}{huggingface.co/gemma-2-2b-it}}

\paragraph{DPO Training Set} The \textit{dpo} and \textit{dpo-ling} training sets are composed of 7.394 (Llama) and 7.246 (Gemma) pairs of preferred and dispreferred responses for the XSUM dataset, and of 6.161 (Llama) and 6.110 (Gemma) for arXiv Abstracts. %
For more details about the selected linguistic features and DPO dataset statistics, we refer the reader to \Cref{app:lingfeats}, and \Cref{app:train-dpo}.

\paragraph{Training Details} We perform DPO fine-tuning of both Llama and Gemma by running a grid search over two hyperparameters: $\beta$ and the learning rate. For all the fine-tuning processes, we leverage LoRA with adapters set to rank $r=32$, as used in the original paper \cite{hu2022lora}, and apply it to all attention layers' weights. For more information, see Appendix \ref{app:training}. 

\paragraph{Human Evaluation} In addition to the evaluation using MGT detectors, we conduct a crowd-based human evaluation. Specifically, for each \textit{(HWT, MGT)} pair, human raters are asked to identify which document is generated by a language model. We evaluate both models using generations obtained before and after the DPO alignment on the XSUM dataset, selecting a random sample of 100 pairs for each configuration. The raters were anonymously recruited among English native speakers via the Prolific online crowd-sourcing platform\footnote{\url{https://www.prolific.com/}}. Five different raters participated in each survey session (each survey comprised 20 pairs, for a total of 25 annotators per model). Further details regarding the annotation process can be found in \Cref{app:human_experiments}.

\section{Results}
\label{sec:results}
To evaluate the effectiveness of our approach, we measure the \textbf{performance drop} of MGT detectors when evaluated on texts generated by adversarially trained models. This is because, if our hypothesis is correct, these detectors, which rely on superficial stylistic clues, should perform worse on the generations of our newly aligned models. \Cref{tab:res-xsum} reports the results obtained by evaluating the pool of MGT detectors presented in \Cref{sec:experimental_setup} before and after the DPO runs. 

\begin{table}[t]
\centering
\input{tables/tpr_at_fpr}
\caption{TPR @ 0.01 FPR and TPR @ 0.05 FPR achieved by the existing supervised detectors when tested on texts generated by Llama and its fine-tunes.}
\label{tab:fpr_at_tpr}
\end{table}

First, the drop in detector accuracy after just one DPO iteration (\textit{dpo-1}, \textit{dpo-1-ling}) indicates that our approach effectively makes MGT detection more challenging. This happens for all the detectors and both data selection strategies: on average MGT detectors lose between 5 (Llama and Gemma on the arXiv Abstract dataset) to 35 percentage points (Llama on XSUM dataset).

The difference in detection performance across datasets highlights the domain sensitivity of MGT detection. Pre-trained detectors (i.e. Mage, Radar, LLM-DetectAIve, and Binoculars) achieve higher accuracy on the XSUM dataset compared to the arXiv Abstracts, as shown by the results highlighted in grey in \Cref{tab:res-xsum}, which report scores for the base LLMs. This discrepancy also affects the effectiveness of our approach, which has a stronger impact on XSUM than on arXiv Abstracts. For instance, Mage, evaluated on the Llama \textit{dpo-1-ling} texts, drops from 76\% to 47\% accuracy on XSUM, while on the arXiv Abstract dataset drops from 77\% to 49\%.

In \cref{tab:res-xsum}, we also report the performance obtained by two detectors explicitly fine-tuned on the two datasets used in this study (i.e., SVM and RoBERTa). In this scenario, the DPO alignment procedure yields MGT harder to detect for all scenarios except for RoBERTa when applied to Gemma on arXiv Abstracts. Additionally, for all four scenarios, the detectors' performance deteriorates more significantly when the DPO fine-tuning leverages random samples rather than samples selected using the scores assigned by the SVM trained on the linguistic profiling. 

To make sense of this behaviour, we ablate the SVM-based detector after removing its top 10 most relevant features, which are those used for selecting the DPO dataset, and find that accuracy remains nearly unchanged (-1\%). This suggests that random sampling texts for the DPO training alters the distribution of a broader range of linguistic features, which helps in the objective of dropping detectors' performance. However, as will be shown in \Cref{sec:intrinsic}, when training on linguistically selected samples (the \textit{dpo-ling} setting), the most relevant features exhibit better alignment to HWT.

For a more detailed analysis of detectors' effectiveness, we report the true positive rates (TPR) at low false positive rates (FPR). \Cref{tab:fpr_at_tpr} presents the TPR @ 1/5\% FPR for the supervised ones, confirming Radar's superior performance on adversarial examples. Specifically, it achieves a TPR of 0.93 at 1\% FPR and retains some effectiveness even when detecting texts generated after adversarial DPO training. A possible explanation for this robustness is that Radar is trained using an adversarial learning setup specifically designed to simulate paraphrasing attacks. This training paradigm may have strengthened the model to handle distributional shifts introduced by our alignment process. In contrast, the second-best detector, Mage, which reaches a TPR of 0.99 at 5\% FPR, experiences a sharp decline below 0.2 when applied to texts generated by models trained with \textit{dpo} or \textit{dpo-ling}. %

Overall, the Gemma model appears more challenging to detect for the tested detectors (but easier for human annotators, as we will highlight in \Cref{sec:human_study}) w.r.t. the Llama model, as highlighted by the results reported in the gray row in \Cref{tab:res-xsum}. Nonetheless, the first iteration of DPO alignment (both \textit{dpo-1} and \textit{dpo-1-ling}) yields consistent drops across both settings, even if smaller w.r.t. those obtained for Llama. Interestingly, in contrast with Llama, whose alignment effects nearly plateau after the first iteration, Gemma's generations appear to benefit from a second alignment step (\emph{dpo-2}, \emph{dpo-2-ling}).

These results demonstrate that even a short DPO run on approximately 7k samples can significantly reduce the accuracy of MGT detectors while remaining fully in-domain and producing grammatical, coherent text. This suggests that the devised approach could be an effective methodology for disrupting current detection patterns, and can be used as a challenging MGT benchmark to assess detectors' capabilities on harder machine-generated text, which can help towards the creation of robust MGT detectors for real-world scenarios.

In the remainder of this article, we provide an in-depth investigation of the linguistic alignment process, focusing on the XSUM dataset before and after the first alignment iteration (\textit{dpo-1-ling}). Details on the arXiv Abstracts dataset are reported in \Cref{app:linguistic_alignment_arxiv}.

\input{tables/manova_distribution}

\subsection{Linguistic Alignment of MGTs}
\label{sec:intrinsic}

To better understand the linguistic alignment between human-written text (HWT) and machine-generated text (MGT), we investigate whether the linguistic profiles of these texts differ significantly after the fine-tuning process (\textit{dpo-1} and \textit{dpo-1-ling}) on the XSUM dataset. To do so, we use a Multivariate ANalysis Of VAriance (MANOVA), which, unlike its univariate version ANOVA, evaluates differences in multivariate mean vectors, allowing us to capture the joint effect of all linguistic features. We report Pillai's Trace, a robust test statistic particularly suited for cases where the assumption of covariance homogeneity may be violated (common in linguistic data due to intercorrelations among features). Higher values of Pillai's Trace indicate stronger multivariate separation between groups. To validate the distinction between human and machine-generated texts, we first compare HWT against MGT outputs. The results (reported in Table \ref{tab:manova}) reveal a Pillai's Trace of 0.7628, suggesting that 76.28\% of the variance is unique to each text type. The value is highly statistically significant ($p < 10^{-5}$), which confirms a strong distinction in linguistic profiles between HWT and baseline MGT, also in line with previous works \cite{krishna-etal-2020-reformulating,esuli2024you}. We also tested the HWT against the outputs of \textit{dpo-1} and \textit{dpo-1-ling} models. Interestingly, \textit{dpo-1-ling} shows a stronger alignment with HWT compared to standard DPO. While DPO yields a Pillai's Trace of 0.7635, which is slightly higher than the non-aligned MGT texts, the \textit{dpo-1-ling} outputs result in a lower Pillai's Trace (0.7137). These findings support that, while \emph{dpo} seems to shift a broader set of linguistic features which helps to fool detectors, \emph{dpo-ling} is more effective in aligning the selected linguistic characteristics of MGT with human writing, resulting in distributions nearer to HWTs. 

To see which linguistic feature gets shifted more during the alignment process, we report in \Cref{tab:xsum-js-iter1} the Jensen-Shannon Divergence (JS) between the linguistic features extracted from the original documents (i.e., HWT) and those generated by the LLMs on XSUM. %
Specifically, we present the results for the top 10 discriminative features selected in the \emph{dpo-1-ling} setting for both models. As shown in the Table, most features are significantly influenced by the \textit{dpo-ling} alignment process, leading to a stronger resemblance to human-written texts compared to the \emph{dpo} strategy. This is especially true for Gemma, for which we notice that most linguistic features, including those used for aligning Llama generations, are better aligned with HWTs after the DPO fine-tuning. Notably, Gemma aligns better with HWTs in morphosyntactic categories, such as the distribution of nouns and adjectives (\textit{upos-dist-*}), as well as word length (\textit{char-per-tok}). In contrast, Llama produces texts with a Type/Token Ratio (TTR) for both forms and lemmas (\textit{ttr-*}), clause length (\textit{avg-token-per-clause}), and numeral distribution (\textit{upos-dist-NUM}) that more closely resemble human-written ones.

\begin{figure}[t!]
    \centering
    \begin{subfigure}{1\linewidth}
        \includegraphics[width=\linewidth]{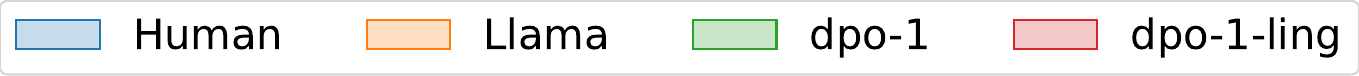}
    \end{subfigure}
    \begin{subfigure}{0.23\textwidth}
        \includegraphics[width=\linewidth]{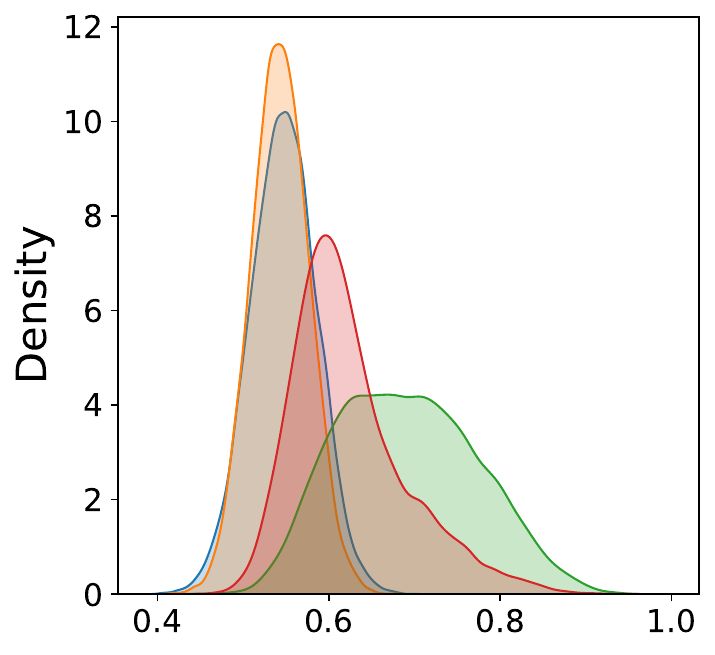}
        \caption{Lexical Density}
        \label{subfig:llama_lingfeates_lexical_density}
    \end{subfigure}
    \begin{subfigure}{0.23\textwidth}
        \includegraphics[width=\linewidth]{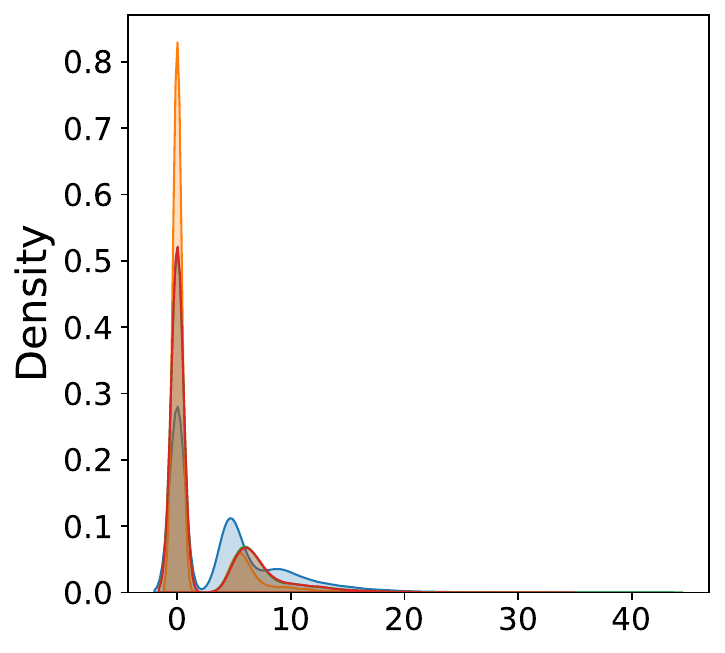}
        \caption{Subj Post}
        \label{subfig:llama_lingfeates_subj_post}
    \end{subfigure}
    
    \begin{subfigure}{0.23\textwidth}
        \includegraphics[width=\linewidth]{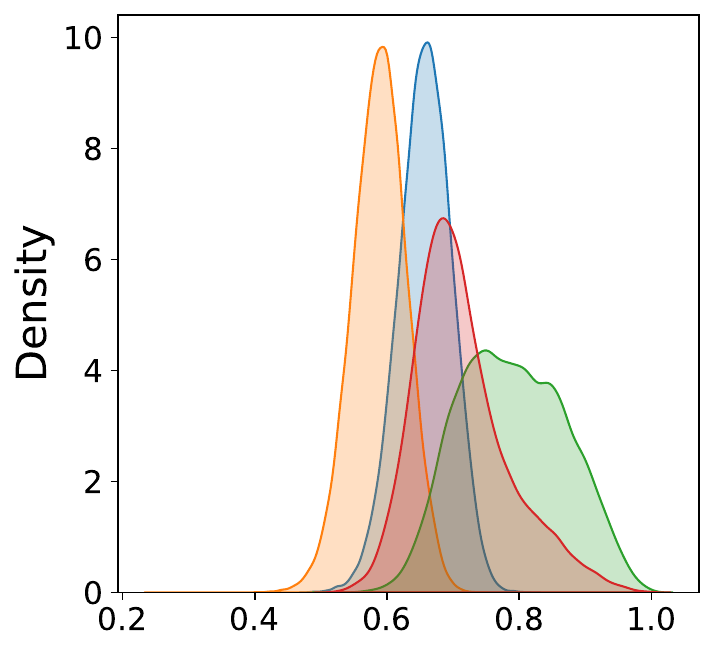}
        \caption{TTR Form @ 200}
        \label{subfig:llama_lingfeats_ttr_form_chunks_200}
    \end{subfigure}
    \begin{subfigure}{0.23\textwidth}
        \includegraphics[width=\linewidth]{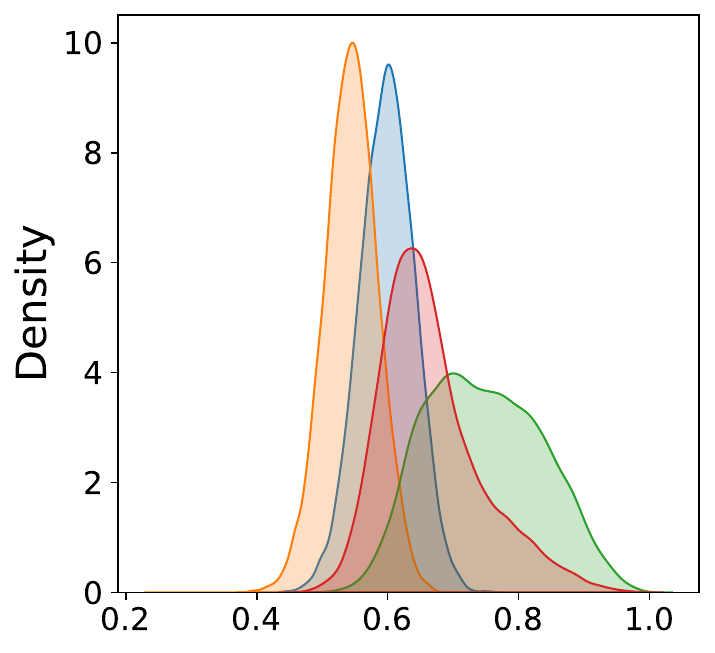}
        \caption{TTR Lemma @ 200}
        \label{subfig:llama_lingfeats_ttr_lemma_chunks_200}
    \end{subfigure}

  \caption{The distribution of selected linguistic features comparing the generations of human, Llama, and the first iteration of our DPO training, in  (\subref{subfig:llama_lingfeates_lexical_density}) Lexical Density, in (\subref{subfig:llama_lingfeates_subj_post}) Subj Post, in (\subref{subfig:llama_lingfeats_ttr_form_chunks_200}) TTR Form Chunks 200 and in (\subref{subfig:llama_lingfeats_ttr_lemma_chunks_200}) TTR Lemma Chunks 200.}
    \label{fig:xsum-llama-lingfeats-iter1}

\end{figure}

\begin{figure}[t!]
    \centering
    \begin{subfigure}{1\linewidth}
        \includegraphics[width=\linewidth]{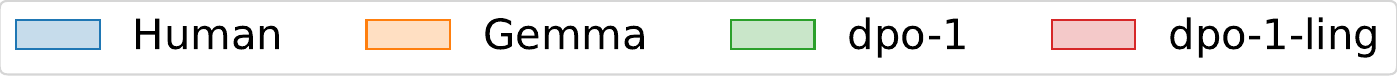}
    \end{subfigure}
    \begin{subfigure}{0.23\textwidth}
        \includegraphics[width=\linewidth]{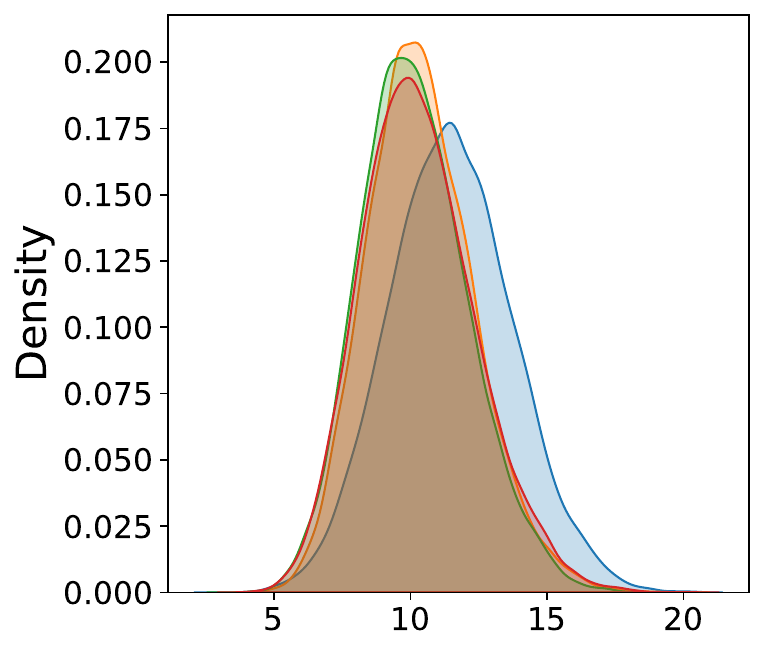}
        \caption{UPoS PUNCT}
        \label{subfig:gemma_lingfeats_upos_dist_PUNCT}
    \end{subfigure}
    \begin{subfigure}{0.23\textwidth}
        \includegraphics[width=\linewidth]{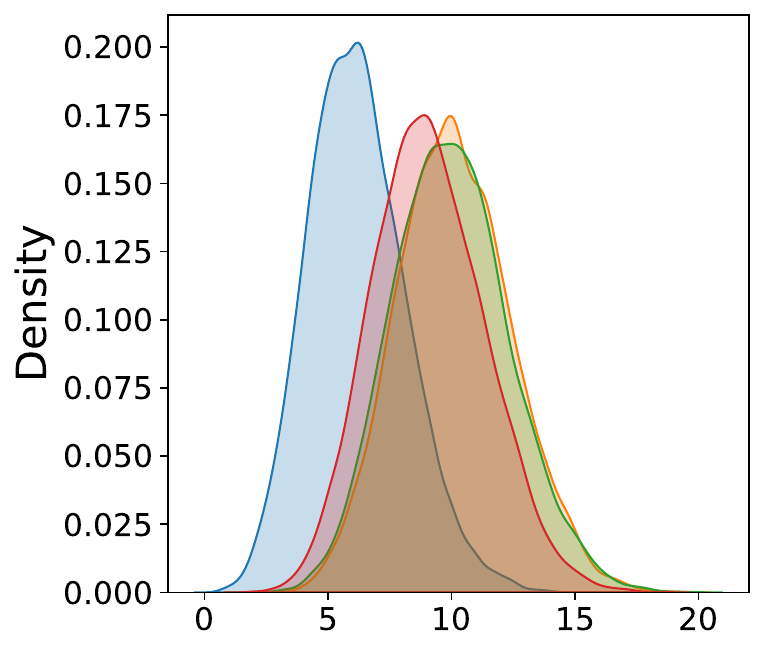}
        \caption{UPoS ADJ}
        \label{subfig:gemma_lingfeats_upos_dist_ADJ}
    \end{subfigure}
    
    \begin{subfigure}{0.23\textwidth}
        \includegraphics[width=\linewidth]{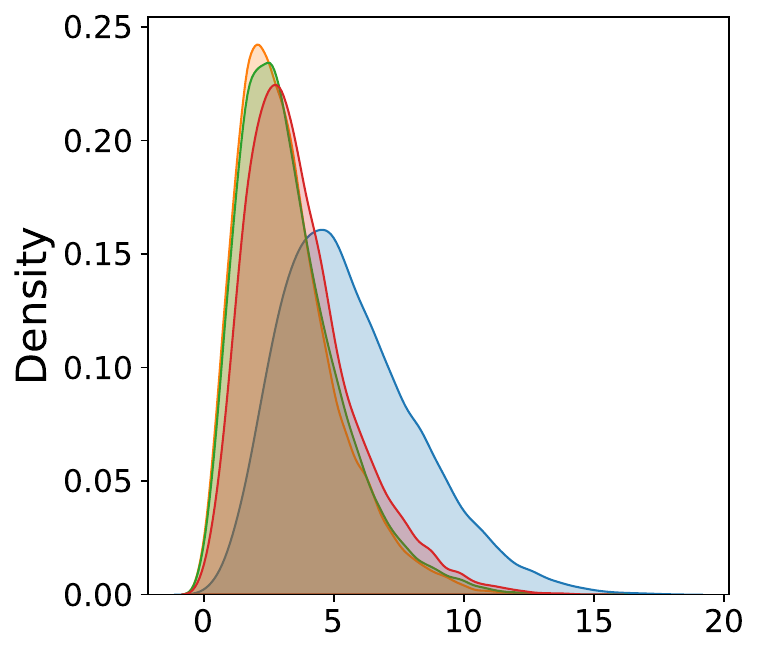}
        \caption{UPoS PRON}
        \label{subfig:gemma_lingfeats_upos_dist_PRON}
    \end{subfigure}
    \begin{subfigure}{0.23\textwidth}
        \includegraphics[width=\linewidth]{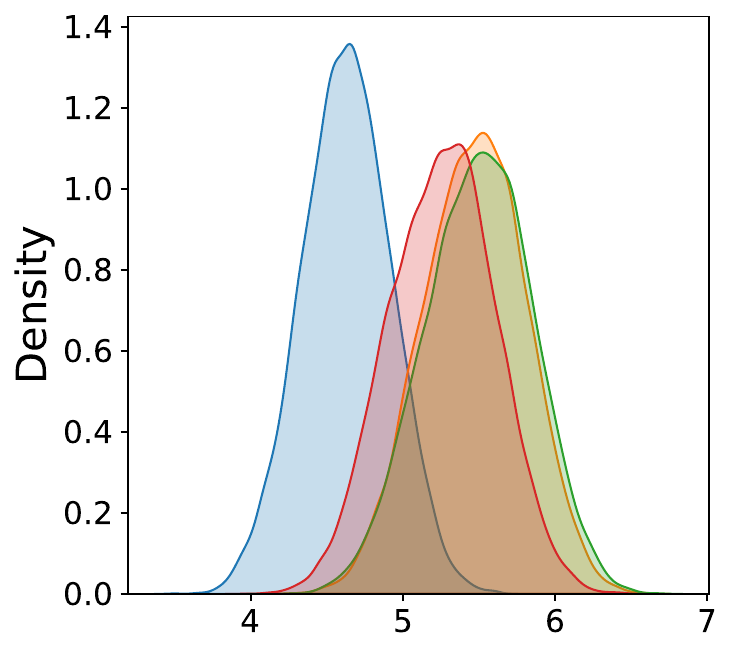}
        \caption{Characters per Token}
        \label{subfig:gemma_lingfeats_char_per_tok}
    \end{subfigure}

  \caption{The distribution of selected linguistic features comparing the generations of human, Gemma, and the first iteration of our DPO training, in  (\subref{subfig:gemma_lingfeats_upos_dist_PUNCT}) UPoS dist PUNCT, in (\subref{subfig:gemma_lingfeats_upos_dist_ADJ}) UPoS dist ADJ, in (\subref{subfig:gemma_lingfeats_upos_dist_PRON}) UPoS dist PRON and in (\subref{subfig:gemma_lingfeats_char_per_tok}) Characters per Token.}
    \label{fig:xsum-gemma-lingfeats-iter1}

\end{figure}

To further analyze the shift in linguistic features before and after DPO alignment, \Cref{fig:xsum-llama-lingfeats-iter1} and \Cref{fig:xsum-gemma-lingfeats-iter1} present the distributions of a subset of selected linguistic features extracted from human-written (\textit{human}) and generated texts (\textit{llama/gemma}, \textit{dpo-1-ling}, \textit{dpo}). As shown in the Figure, and consistent with \Cref{tab:xsum-js-iter1}, the linguistically-guided alignment (\textit{dpo-1-ling}) brings the distribution of TTR-related features for Llama and of Part-of-Speeches (POS) for Gemma closer to that of human-written documents. Interestingly, for features where human-written and generated texts were already well aligned before the DPO fine-tuning, e.g. lexical density\footnote{The Lexical Density is computed as the ratio between lexical words and all the tokens in a document.} for the Gemma model, the alignment process tends shift them away, leading to a slight misalignment with HWTs. However, even in these cases, the \textit{dpo-ling} strategy still produces outputs that remain closer to human-written texts compared to \textit{dpo}, highlighting the overall effectiveness of the linguistically-guided alignment in tuning the model to generate texts that are stylistically more "human". 

\begin{table*}[t!]
\centering
\input{tables/all_jensen}
\caption{Jensen-Shannon divergence of linguistic features between HWTs and MGTs %
by the base LLMs and our adversarial fine-tuned models on XSUM. In \textbf{bold} the lowest value among models sharing the base model.}
\label{tab:xsum-js-iter1}
\end{table*}

\subsection{Comparison with Human Raters}
\label{sec:human_study}

\begin{figure}[t!]
    \centering
    \includegraphics[width=0.9\linewidth]{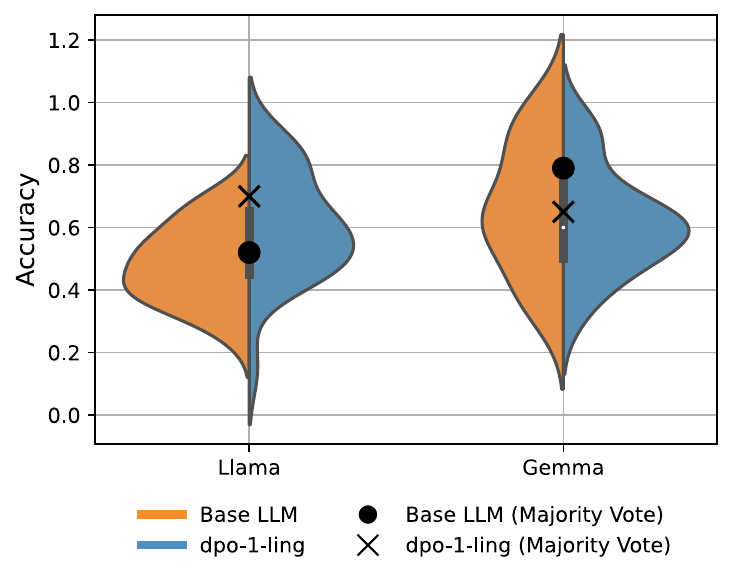}
    \caption{Distribution of accuracy scores among all the annotators for both models (before and after the DPO alignment). Accuracy scores computed with the majority vote are also reported.}
    \label{fig:annotators_plot}
\end{figure}

As discussed in \Cref{sec:experimental_setup}, we also inspect the effect of DPO alignment on the ability of human raters to recognize MGTs. We conducted this evaluation for the Llama and Gemma models before and after the first iteration of DPO alignment on the XSUM dataset. \Cref{fig:annotators_plot} reports the distribution of accuracy scores obtained for each annotator individually, as well as the scores achieved with a majority voting system (across 5 annotators). We observe that the two models exhibit distinct behaviors after the DPO alignment. Specifically, the Llama model becomes more easily detectable by human raters post-alignment, while the opposite trend is observed for Gemma, whose generations become harder to recognize. Nevertheless, it is important to notice that for all the tested models we obtained low agreement scores, ranging between 0.06 and 0.10 (as measured by Fleiss' Kappa\footnote{Agreement are reported in \Cref{app:human_experiments}.}), thus highlighting the difficulty of the task for human annotators. Focusing instead on the scores obtained by individual annotators, although there is a difference between models and before and after the DPO alignment, the majority of the accuracies are distributed between 0.40 and 0.60. This suggests that the raters are often close to random guessing, thus further reinforcing the difficulty of distinguishing machine-generated from human-written text.

\section{Discussion and Conclusion}
\label{sec:conclusions}

In this work, we develop a pipeline to generate synthetic texts that more closely align with the style of human-written texts, thereby making them more challenging for automatic MGT detectors and suitable to be used as a robust benchmark. To do this, we select paired human- and machine-written texts and use Direct Preference Optimization (DPO) to align LLMs' writing style to those of humans in specific domains, such as news and scientific writing. 
The result of our study is twofold: 

\begin{itemize}
    \item we provide a framework to generate more challenging MGT data that can be used to evaluate the robustness and reliability of existing MGT detectors, and hopefully to create new, more robust ones;
    \item we highlight the limitations and flaws of current MGT detection approaches (i.e. the reliance on the linguistic shortcuts inferable from the generations) and present an in-depth investigation of how the alignment step changes the text produced by the LLMs.
\end{itemize}

\paragraph{MGT Detectors Limitations:} We show that texts generated by our newly aligned models exhibit stronger human-like properties, leading to a significant drop in detector accuracy, up to 60\%. 
This shows that, as previously demonstrated by \citet{doughman-etal-2025-exploring}, MGT detectors rely on superficial stylistic clues and thus can be easily fooled by this strategy. Future work should focus on making MGT detectors more robust to these types of attacks, similar to how RADAR was made more robust to paraphrasing attacks, which likely improved its performance against our linguistically informed adversarial attacks.

\paragraph{Linguistic Analysis of MGT:} We analyze the linguistic properties of machine-generated texts before and after adversarial fine-tuning and compare them to human-written texts. We find that using randomly sampled couples of MGT and HWT (\emph{dpo-1}) provides a broader range of linguistic features that get shifted towards HWT style. This seems to be the most effective technique against most of the detectors (RADAR, the more robust one, being one of the exceptions) if the objective is to create more challenging texts for detectors. However, if we want feature-specific alignment, we find that using our linguistically-informed data selection technique (\emph{dpo-1-ling}) for DPO creates texts that more closely align with HWT w.r.t. the selected linguistic features.

We also find that doing more iterations of our pipeline, to further align the models towards HWT (\emph{dpo-2}, \emph{dpo-2-ling}), can be beneficial, depending on the generating model (e.g., Gemma benefits more than Llama from this) and on the detector. However, it must be noted that most of the detectors experience significant drops in performance after just one iteration.

Finally, our human evaluation highlights how complex the task of distinguishing between HWT and MGT is from the perspective of human evaluators. 
Notably, we find that after the alignment process, the performance of human annotators is not significantly affected, remaining mostly random guessing. This result further corroborates how our alignment process doesn't just make the text harder for the detectors, but also remains just as hard for humans, speaking to the quality of the texts.

We believe that our proposed pipeline, along with our findings, will facilitate the development of more challenging and robust benchmarks for MGT detection, ultimately contributing to the responsible deployment of generative AI.

\section{Related Works}
\label{sec:sota}
The need to mitigate the risks of spreading machine-generated content has driven the literature on Machine Generated Text (MGT) Detection along several research directions.

\subsection{MGT Detection Methodologies} 
A straightforward way to mitigate the risk of machine-generated text being mistaken for human-written is to train supervised detectors on large-scale datasets to distinguish between the two.

Among these, \citet{li-etal-2024-mage} introduce MAGE (MAchine-GEnerated text detection) and study the resilience of detectors to several train-test distribution shifts. %
\citet{hu2023radar} propose RADAR, a detector designed to be robust against paraphrased MGTs, addressing a known limitation of general-purpose detectors \citep{uchendu2023attribution}. LLM-DetectAIve \citep{abassy-etal-2024-llm}
classifies texts into four categories based on whether they were initially written by a human or a machine and later edited by the other. Earlier studies also explored traditional authorship attribution methods, treating LLMs as authors \citep{uchendu_authorship_2020}.

In addition to supervised approaches, in the past few years, several works have focused on developing statistical methods, building on the key idea that MGTs are generally more likely under LLM probability distributions than human-written text. One of the earliest studies in this direction is Detect-GPT \citep{mitchell_detectgpt_2023}. An extension of this work, DetectLLM \citep{su-etal-2023-detectllm}, further examines the computational costs of statistical detectors. Binoculars \citep{hans2024spotting} improves detection accuracy by combining likelihood estimates from two different LLMs. %

\subsection{MGT Detection Benchmarks} 
The systematic development of detectors has also led the way for the creation of various benchmarks and studies focused on dataset development. One of the earliest examples is M4 \citep{wang-etal-2024-m4}, a dataset consisting of paired human-written and machine-generated texts on the same topics, covering multiple domains and languages. \citet{wang2024m4gtbench} introduce an extension of M4, i.e. M4GT, by formulating three tasks and defining a new formulation of detecting the changing point from human-written to machine-generated. Similarly, \citet{dugan2024raid} presents RAID, 
spanning different models, domains, adversarial attacks, and decoding strategies. Focusing instead on the social-media domain, \citet{macko2024multisocialmultilingualbenchmarkmachinegenerated} built MultiSocial, a multilingual dataset for MGT detection derived from 5 different social media platforms.

\subsection{MGT Detection Limitations}

Although existing benchmarks are comprehensive, they can not be used as proxies of real world scenarios, as suggested by the results of multiple shared tasks on MGT detection \citep{wang-etal-2024-semeval-2024, dugan2025genaicontentdetectiontask, wang-etal-2025-genai}, where winning participants achieve near perfect accuracy. 
However, this high performance does not necessarily extend to real-world scenarios, particularly when evaluating randomly sourced texts -- a limitation also acknowledged by professional MGT detection providers\footnote{\href{https://gptzero.me/}{https://gptzero.me/}} with dangerous results in real applications\footnote{\href{https://eu.usatoday.com/story/news/education/2023/04/12/how-ai-detection-tool-spawned-false-cheating-case-uc-davis/11600777002/}{https://eu.usatoday.com/}}.
Indeed, \citet{Puccetti_2024} show that models fine-tuned on non-English languages can be more challenging to detect, while \citet{doughman-etal-2025-exploring} identify shallow linguistic cues that MGT detectors rely on, such as punctuation patterns and average word length. Additionally, they highlight that dataset shift has an important effect on the performance of MGT detectors: when the domain under analysis is absent from the detector's training set, performance significantly drops.

\subsection{Linguistic Profiling}
As pointed out by \citet{doughman-etal-2025-exploring}, MGT detectors can rely on linguistic clues to identify machine-generated texts. Therefore, applying \textit{linguistic profiling} techniques -- which analyze stylistic, syntactic, and lexical characteristics of a text -- can offer a systematic way to capture such clues. \textit{Linguistic profiling} is a
NLP-based methodology in which a large set of linguistically motivated features automatically extracted from annotated texts are used to obtain a
vector-based representation of it. Such representations can be then exploited for e.g. comparing and analyzing different textual genres \cite{van-halteren-2004-linguistic}, measuring natural language complexity \cite{thompson-etal-2014-computational} or profiling the linguistic knowledge implicitly encoded in the internal representations of LMs \cite{miaschi-etal-2020-linguistic}. Different tools exist today to perform linguistic profiling, making use of different types of features \cite{eder2011stylometry,lee-etal-2021-pushing}. Among these, we can mention ProfilingUD \cite{brunato-etal-2020-profiling}, a tool that allows the extraction of more than 130 linguistic properties 
based on the UD formalism \cite{10.1162/coli_a_00402}.

\section*{Limitations}

This work investigates weaknesses in models trained to detect MGT. We fine-tune existing LLMs on a limited set of examples to better mimic human writing style, significantly reducing the accuracy of current MGT detectors. Additionally, we analyze the linguistic properties of texts generated by the original LLMs and those produced after our linguistically-informed DPO training. While we demonstrate a reduction in the performance of specific detectors, the extent of accuracy loss varies across models and domains. Expanding our evaluation to a broader set of models, particularly larger ones, would enhance the robustness of our findings.

Regarding datasets, we focus on two key domain -- news and scientific writing. However, the risks associated with undisclosed MGT extend beyond these domains. Therefore, future work could explore additional domains where such use may be particularly problematic.

Finally, we conduct a human evaluation to measure how well humans can detect texts generated by LLMs. We interpret raters performance, which stays largely unchanged, as a proxy measure of the fact that after dpo fine-tuning our models still generate grammatical and coherent texts. However, our study is limited in scope and not primarily focused on assessing the coherence and the grammaticality of generated documents. Therefore, it would be beneficial, in future work, to incorporate targeted assessment of fluency and readability to better evaluate the quality of the generated texts.

\section*{Ethics Statement}

We acknowledge that our approach, which fine-tunes Language Models to generate text more similar to human-written content, could be exploited to develop technologies aimed at evading MGT detectors. In particular, malicious actors could adapt our method to generate synthetic content that is more difficult to distinguish from human-authored text, potentially facilitating misinformation or manipulation. We argue that, while malicious actors would keep pursuing their goals regardless of the community efforts, our work aims to expose the vulnerabilities of current detection systems and highlight their reliance on linguistic shortcuts, ultimately contributing to the development of more robust and generalizable MGT detection methods. By openly sharing our findings and dataset we provide the research community with the necessary tools to improve detection strategies and mitigate the risks associated with increasingly sophisticated AI-generated content. However, to mitigate malicious use of our findings we will release code and models upon request. 

\section*{Acknowledgements}
Andrea Pedrotti is fully funded by the European Union - NextGenerationEU through PNRR (CUP B53C22001760006) ``SoBigData.it: Strengthening the Italian RI for Social Mining and Big Data Analytics'' (SoBigData.it). Giovanni Puccetti is fully funded by the Italian Ministry of University and Research under the PNRR project ITSERR (CUP B53C22001770006). This project is also funded by the project XAI-CARE-PNRR-MAD-2022-12376692 under the NRRP MUR program funded by the NextGenerationEU and the PNRR MUR project \href{https://fondazione-fair.it/}{PE0000013-FAIR}. This project is also partially funded by the project ``Word Embeddings: From Cognitive Linguistics to Language Engineering, and Back'' (WEMB), funded by the Italian Ministry of University and Research (MUR) under the PRIN 2022 funding scheme (CUP B53D23013050006).

\bibliography{custom, anthology_0.bib, anthology_1.bib}

\appendix
\input{appendix}

\end{document}

%% file: tables/all_results_horizontal.tex
\begin{tabular}{lcccccccccccc}

        Detector $\rightarrow$  & \rotatebox{90}{Mage}  & \rotatebox{90}{Radar} & \rotatebox{90}{LLM-DetectAIve}  & \rotatebox{90}{Binoculars}  & \rotatebox{90}{SVM\textsuperscript{\textdagger}} & \rotatebox{90}{RoBERTa\textsuperscript{\textdagger}}  & \rotatebox{90}{Mage}  & \rotatebox{90}{Radar} & \rotatebox{90}{LLM-DetectAIve}  & \rotatebox{90}{Binoculars}  & \rotatebox{90}{SVM\textsuperscript{\textdagger}} & \rotatebox{90}{RoBERTa\textsuperscript{\textdagger}}\\
        \cline{2-13}
        \multicolumn{1}{l|}{Generator $\downarrow$} &  \multicolumn{6}{c|}{\textbf{XSUM}} &   \multicolumn{6}{c|}{\textbf{arXiv Abstracts}} \\
        \hline
        
        \rowcolor{gray!30} \multicolumn{1}{l|}{\textbf{Llama}}                     & 0.76  & 0.94  & 0.72  & 0.99  & 0.94  & \multicolumn{1}{c|}{1.00}                        & 0.77  & 0.38  & 0.50  & 0.79  & 0.96  & \multicolumn{1}{c|}{1.00}      \\
        \hline
          \multicolumn{1}{l|}{dpo-1}               & \textbf{0.40}  & 0.79  & 0.53   & \textbf{0.33}  & 0.69 & \multicolumn{1}{c|}{0.52}  &  0.65  & \textbf{0.37}  & 0.46  & 0.61  & 0.95  & \multicolumn{1}{c|}{0.99}      \\
         \multicolumn{1}{l|}{dpo-1-ling}        & 0.47  & \textbf{0.58}  & 0.54  & 0.38  & 0.80  & \multicolumn{1}{c|}{0.75}         & 0.49  & 0.40  & 0.45  & 0.41  & 0.92  & \multicolumn{1}{c|}{0.97} \\

         \multicolumn{1}{l|}{dpo-2}               & 0.44  & 0.87  & 0.55  & \textbf{0.33}  & \textbf{0.64}  & \multicolumn{1}{c|}{\textbf{0.50}}  &  0.53  & 0.41  & \textbf{0.43}  & 0.43  & 0.94  & \multicolumn{1}{c|}{0.98}  \\
         \multicolumn{1}{l|}{dpo-2-ling}        & 0.42  & 0.62  & \textbf{0.52}  & \textbf{0.33}  & 0.77  & \multicolumn{1}{c|}{0.68}          & \textbf{0.48}  & 0.41  & 0.44  & \textbf{0.39}  & \textbf{0.92}    & \multicolumn{1}{c|}{0.96}   \\

        \hline
        \rowcolor{gray!30} \multicolumn{1}{l|}{\textbf{Gemma}}                     & 0.71  & 0.70  & 0.68  & 0.37 & 0.97  & \multicolumn{1}{c|}{1.00}                       & 0.76  & 0.38  & 0.54  & 0.58  & 0.99  & \multicolumn{1}{c|}{1.00}     \\
        \hline
         \multicolumn{1}{l|}{dpo-1}               & 0.65  & 0.60  & 0.64  & \textbf{0.34} & \textbf{0.70}  & \multicolumn{1}{c|}{0.99}             & 0.70  & \textbf{0.37}  & 0.53  & 0.45  & 0.93  & \multicolumn{1}{c|}{0.99}   \\
         \multicolumn{1}{l|}{dpo-1-ling}        & 0.67  & 0.61  & 0.66  & 0.35   & 0.91  & \multicolumn{1}{c|}{0.99}  &  0.75  & 0.40  & 0.54  & \textbf{0.39}    & 0.96  &  \multicolumn{1}{c|}{0.99}  \\
         \multicolumn{1}{l|}{dpo-2}               & \textbf{0.64}  & 0.58  & \textbf{0.63}  & \textbf{0.34} & 0.96  & \multicolumn{1}{c|}{0.99} & \textbf{0.63}  & \textbf{0.37}  & \textbf{0.52}  & 0.40  & \textbf{0.83}  & \multicolumn{1}{c|}{\textbf{0.92}}  \\
         \multicolumn{1}{l|}{dpo-2-ling}        & 0.65  & \textbf{0.56} & 0.64  & 0.35  & 0.83  & \multicolumn{1}{c|}{0.99}        & 0.70  & 0.38  & 0.52  & 0.47  & 0.87  & \multicolumn{1}{c|}{0.96} \\
        \hline

    \end{tabular}

%% file: tables/tpr_at_fpr.tex
\adjustbox{max width=\linewidth}{
\begin{tabular}{lccc}
\toprule
  & Llama & dpo-1 & dpo-1-ling \\
\midrule
\multicolumn{4}{c}{TPR @ 0.01 FPR} \\
\midrule
 DetectAIve & 0.312 & 0.001 & 0.017 \\
 Mage       & 0.054 & 0.014 & 0.066 \\
 Radar      & 0.932 & 0.620 & 0.324 \\
\midrule
\multicolumn{4}{c}{TPR @ 0.05 FPR} \\
\midrule
DetectAIve & 0.428 & 0.010 & 0.044 \\
 Mage       & 0.997 & 0.057 & 0.176 \\
 Radar      & 0.995 & 0.762 & 0.571 \\
\bottomrule
\end{tabular}
}

%% file: tables/manova_distribution.tex
\begin{table}[t!]
\centering
\begin{tabular}{lcc}
\hline
\textbf{Comparison} & \textbf{Pillai's Trace} & \textbf{p-Value} \\
\hline
HWT vs Base LLM       & 0.7628 & $< 10^{-5}$ \\
HWT vs \textit{dpo-1}       & 0.7635 & $< 10^{-5}$ \\
HWT vs \textit{dpo-1-ling}  & \textbf{0.7137} & $< 10^{-5}$ \\
\hline
\end{tabular}
\caption{Pillai's Trace and p-values for the MANOVA test computed between the linguistic features extracted from HWT and MGT before and after the first step of both DPO and DPO-ling on the XSUM dataset.}
\label{tab:manova}
\end{table}

%% file: tables/all_jensen.tex
\adjustbox{max width=\linewidth}{
\begin{tabular}{lcccccccccccccccccc}
 Model & \rotatebox{90}{\textbf{subj-post}} & \rotatebox{90}{\textbf{upos-dist-PROPN}} & \rotatebox{90}{\textbf{lexical-density}} & \rotatebox{90}{\textbf{n-tokens}} & \rotatebox{90}{\textbf{ttr-lemma-chunks-200}} & \rotatebox{90}{\textbf{avg-token-per-clause}} & \rotatebox{90}{\textbf{char-per-tok}} & \rotatebox{90}{\textbf{ttr-form-chunks-100}} & \rotatebox{90}{\textbf{ttr-form-chunks-200}} & \rotatebox{90}{\textbf{upos-dist-PUNCT}} & \rotatebox{90}{\textbf{n-sentences}} & \rotatebox{90}{\textbf{tokens-per-sent}} & \rotatebox{90}{\textbf{upos-dist-PRON}} & \rotatebox{90}{\textbf{ttr-lemma-chunks-100}} & \rotatebox{90}{\textbf{upos-dist-ADJ}} & \rotatebox{90}{\textbf{upos-dist-NOUN}} & \rotatebox{90}{\textbf{verbs-form-dist-Ger}} & \rotatebox{90}{\textbf{upos-dist-NUM}} \\
\midrule
Llama & .266 & .261 & \textbf{.103} & .207 & .431 & .149 & \textbf{.274} & .405 & .509 & \textbf{.230} & \textbf{.374} & \textbf{.427} & \textbf{.259} & .333 & \textbf{.241} & \textbf{.341} & \textbf{.349} & .238 \\
dpo-1-ling & .213 & .134 & .447 & \textbf{.118} & \textbf{.343} & \textbf{.065} & .561 & \textbf{.052} & \textbf{.318} & .383 & .464 & .489 & .312 & \textbf{.068} & .403 & .467 & .433 & \textbf{.224} \\
dpo-1 & \textbf{.209} & \textbf{.105} & .635 & .150 & .537 & .259 & .663 & .213 & .536 & .569 & .551 & .568 & .376 & .266 & .480 & .579 & .459 & .258 \\
\midrule
Gemma & .249 & .283 & .508 & \textbf{.181} & .289 & .079 & .681 & .061 & .231 & \textbf{.215} & .284 & .331 & .346 & .064 & .556 & .494 & .573 & .481 \\
dpo-1-ling & \textbf{.194} & \textbf{.262} & \textbf{.369} & .202 & \textbf{.258} & .071 & \textbf{.565} & .071 & \textbf{.225} & .216 & \textbf{.216} & .270 & \textbf{.274} & \textbf{.055} & \textbf{.445} & \textbf{.427} & \textbf{.453} & \textbf{.421} \\
dpo-1 & .236 & .276 & .519 & .242 & .336 & \textbf{.061} & .679 & \textbf{.055} & .272 & .227 & .259 & \textbf{.263} & .321 & .119 & .535 & .493 & .563 & .493 \\
\bottomrule
\end{tabular}
}

%% file: appendix.tex
\section{Detectors}
\label{sec:detectors}
In the paper we use 3 existing supervised detectors, 2 custom supervised detectors and 1 statistical detector, see \Cref{tab:detectors} for details about each of them:

\begin{table}[ht]
    \centering
    \adjustbox{max width=\linewidth}{
    \begin{tabular}{ccc}
    \toprule
         Detector & Type & Reference \\
    \midrule
        Mage & Existing Supervised & \citep{li-etal-2024-mage} \\
        Radar & Existing Supervised & \citep{hu2023radar} \\
        DetectAIve & Existing Supervised & \citep{abassy-etal-2024-llm} \\
    \midrule
        SVM & Custom Supervised & -- \\
        RoBERTa & Custom Supervised & -- \\
    \midrule
        Binoculars & Statistical & \citep{hans2024spotting} \\
    \bottomrule
    \end{tabular}}
    \caption{Summary Table of the detectors used throughout this work. The column Type denotes whether the MGT detectors was explicitly trained on the evaluation dataset.}
    \label{tab:detectors}
\end{table}

\begin{table*}[t!]
\scriptsize
    \centering
 \input{tables/lingfeats_explained}
 \caption{Linguistic features extracted with ProflingUD.}
 \label{tab:lingfeat_explained}
\end{table*}

\begin{table*}[ht]
    \centering
    \input{tables/iter_1_feats}
    \caption{First iteration, \textit{dpo-1-ling}, target features.}
    \label{tab:feats-xsum-iter-1}
\end{table*}

\begin{table*}[ht]
    \centering
    \input{tables/iter_2_feats}
    \caption{Second iteration, \textit{dpo-2-ling}, target features.}
    \label{tab:feats-xsum-iter-2}
\end{table*}

\section{Linguistic Features}\label{app:lingfeats}

For a detailed description of the linguistic features we use throughout the work, to select DPO training data and to measure how closely MGTs and HWTs align, see \Cref{tab:lingfeat_explained}.

The linguistic features used for the first and second iteration of the DPO alignment are reported in \Cref{tab:feats-xsum-iter-1} and \Cref{tab:feats-xsum-iter-2}.

\section{DPO Training Details}\label{app:training}
We perform hyperparameter optimization over the following values:
$\beta = \{0.1, 0.5, 1.0\}$ and learning rate: $lr = \{5e^{-7}, 5e^{-6}\}$.

\subsection{Iterative DPO Dataset Construction}
\label{app:train-dpo}

In \cref{app:tab:xsumprompt} and \cref{app:tab:m4absprompt}, we report the prompt used throughout our experimentation.

\begin{table}[ht]
    \centering
    \input{tables/xsum-prompt}
\caption{XSUM prompts.}
\label{app:tab:xsumprompt}
\end{table}

\begin{table}[ht]
    \centering
    \input{tables/m4abs-prompt}
\caption{arXiv Abstract prompts.}
\label{app:tab:m4absprompt}
\end{table}

To construct the preference dataset $\mathcal{D}_{DPO}$ (step $6$ of Algorithm \ref{alg:cap}) in the \emph{dpo} setting, we randomly select couples of (HWT, MGT) and tag the HWT as the preferred option. For the successive iterations, for the couples (HWT, MGT) we select MGT generated by the model aligned during the first iteration. However, we avoid selecting the same HWT twice during multiple iterations. 
 
For the first iteration of \emph{dpo-ling}, we train an SVM classifier to distinguish between HWT and MGT using the profiling feature of the texts as inputs. This is done by constructing a balanced dataset $\mathcal{D}_{par}$ containing HWT and MGT texts (step $4$) and using it as the training set for the SVM.
Then, we take the top ten features with the highest absolute coefficients for the SVM classification, and, for each of these features, we take the top-$k$ pairs $(s_{HWT}, s_{MGT})| s_{HWT}, s_{MGT} \in \mathcal{D}_{par}$ that maximize the feature distance between the two texts:
$$
abs(s_{HWT}[feature] - s_{MGT}[feature])
$$
Then, we label the HWT as the preferred one, while the MGT is labelled as the dispreferred\footnote{However when the SVM classifies the MGT as human-written and the HWT as machine-generated we label the MGT as the preferred one.}, obtaining $\mathcal{D}_{DPO}$ which is used to align the model towards HWT linguistic features.

For the subsequent iterations, we do the same, but the SVM dataset $\mathcal{D}'_{par}$ is built with MGT generated by the model that was aligned through DPO using $\mathcal{D}_{DPO}$, i.e. $\mathcal{M}'$. We also use a subsample of all the sentences where each HWT sentence that was already selected for a previous iteration is not selected $\{(s_{HWT}, s_{MGT}) | s_{HWT}\in\mathcal{D}'_{par}, s_{HWT}\notin \mathcal{D}_{DPO}\}$, and where the feature distance between the texts for the features selected in the previous iteration is near 0. So, $\forall s \in \mathcal{D}':$
$$
s_{HWT}[feature_{\mathcal{D}_{par}}]-s_{MGT}[feature_{\mathcal{D}_{par}}] = 0 \pm \epsilon
$$

This is done so that the sentences we select for the iteration are representative mostly of the current top ten features, and not of the features selected in the previous iterations. We also ensure that the same features are not selected across multiple iterations. Once we obtain $\mathcal{D}'_{DPO}$ we can further align the model towards HWT and continue the process iteratively, if desired. 

For our experiments, we set $k = 1000$ and $\epsilon = 0.1$ for XSUM and $\epsilon = 0.2$ for the arXiv Abstracts dataset. We set a higher value of $\epsilon$ for the arXiv Abstract dataset to obtain a sufficient amount of training documents since the dataset is smaller than XSUM. We keep the dimensions of $D_{DPO}$ equal between the two settings (\emph{dpo}, \emph{dpo-ling}) to obtain comparable results across the two settings, both in terms of impact on the MGT detectors' performance and the quality of the alignment of the textual features.
For the first iteration of the XSUM dataset, we have 7.394 couples for the llama model and 7.246 couples for the Gemma model.
For the second iteration, we obtain an XSUM dataset consisting of 1.583 pairs for Llama, and 3.530 for Gemma. For the arXiv Abstract, we end up with 6.161 paired documents for Llama, and 6.110 for Gemma for the first iteration, and 1.225 and 1.510 for the second iteration, respectively.

\subsection{LoRA}
\begin{python}
LoraConfig(
    r=32,
    lora_alpha=16,
    target_modules=[
        "q_proj",
        "k_proj",
        "v_proj",
        "o_proj",
        "gate_proj",
        "up_proj",
        "down_proj",
        ],
    bias="none",
    lora_dropout=0.05,
    task_type="CAUSAL_LM")
\end{python}

\subsection{Sampling Parameters}

\begin{python}
SamplingParams(
    max_tokens=512,
    min_tokens=256,
    frequency_penalty=0.0,
    repetition_penalty= 1.0,
    temperature=1.0,
    top_p=1.0)
\end{python}

\section{Human Annotation Details}
\label{app:human_experiments}

\begin{table}[ht]
\centering
\small
\begin{tabular}{lrr}
\toprule
\textbf{Generator} & \textbf{Accuracy} & \textbf{Agreement} \\
\midrule
Llama & 0.52 & 0.07 \\
LLama-dpo-1-ling & 0.70 & 0.06 \\
Gemma & 0.79 & 0.10 \\
Gemma-dpo-1-ling & 0.65 & 0.08 \\
\bottomrule
\end{tabular}
\caption{Results of the human evaluation (majority vote across 5 annotators). The agreement is computed using Fleiss' Kappa.}
\label{tab:human_evaluation}
\end{table}

\begin{table*}[ht]
\centering
\input{tables/all_jensen_m4abs}
\caption{Jensen-Shannon divergence of linguistic features between HWTs and MGTs %
by the original LLMs and our adversarial fine-tunes when generating texts from the arXiv Abstract datasets. In \textbf{bold} the lowest value among models sharing the base model.}
\label{tab:m4abs-js-iter1}
\end{table*}

\begin{figure}[ht]
    \centering
    \begin{subfigure}{1\linewidth}
        \includegraphics[width=\linewidth]{figures/legend_lingfeats_llama.pdf}
    \end{subfigure}
    \begin{subfigure}{0.23\textwidth}
        \includegraphics[width=\linewidth]{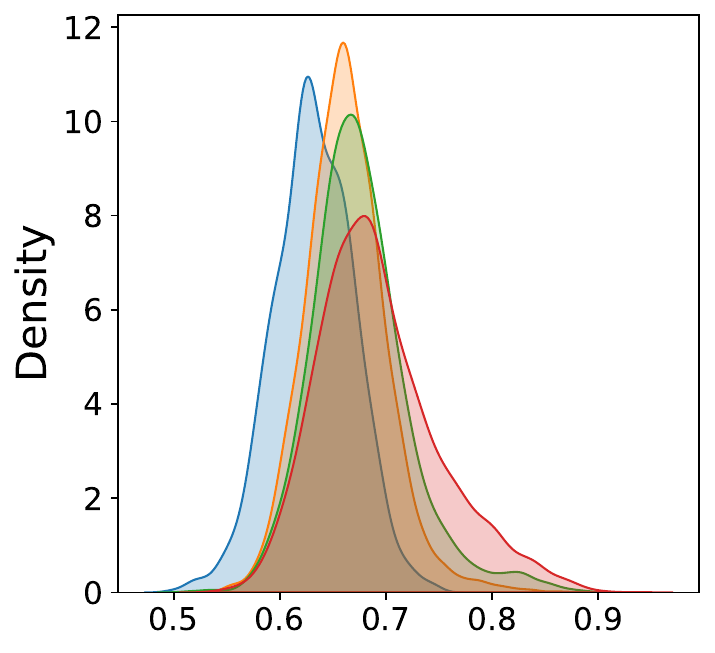}
        \caption{Lexical Density}
        \label{subfig:m4abs_llama_lingfeates_lexical_density}
    \end{subfigure}
    \begin{subfigure}{0.23\textwidth}
        \includegraphics[width=\linewidth]{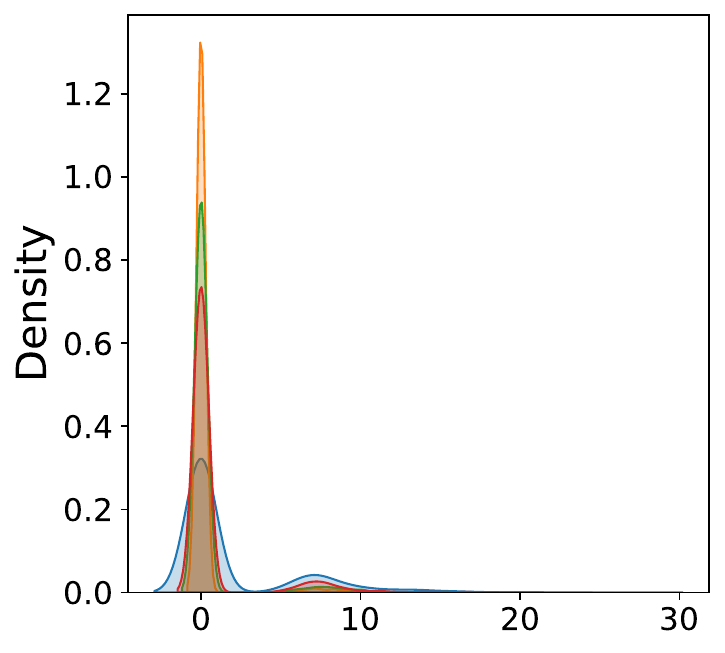}
        \caption{Subj Post}
        \label{subfig:m4abs_llama_lingfeates_subj_post}
    \end{subfigure}
    
    \begin{subfigure}{0.23\textwidth}
        \includegraphics[width=\linewidth]{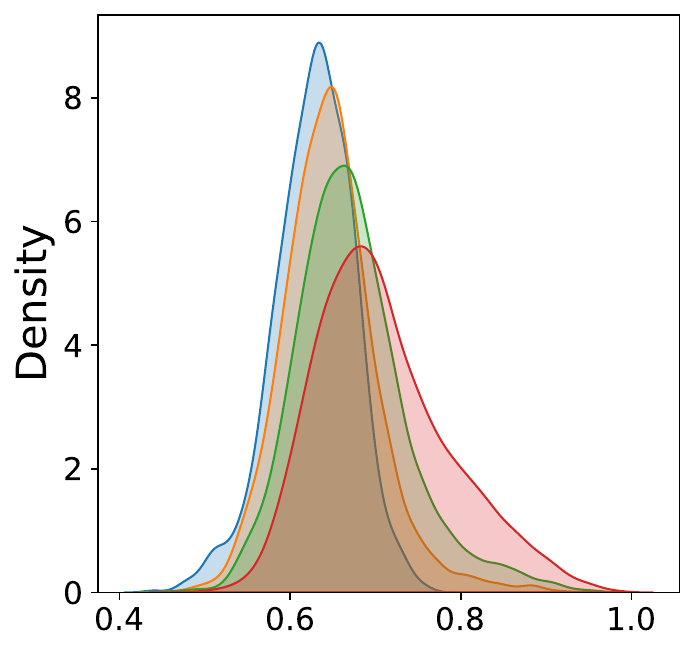}
        \caption{TTR Form @ 200}
        \label{subfig:m4abs_llama_lingfeats_ttr_form_chunks_200}
    \end{subfigure}
    \begin{subfigure}{0.23\textwidth}
        \includegraphics[width=\linewidth]{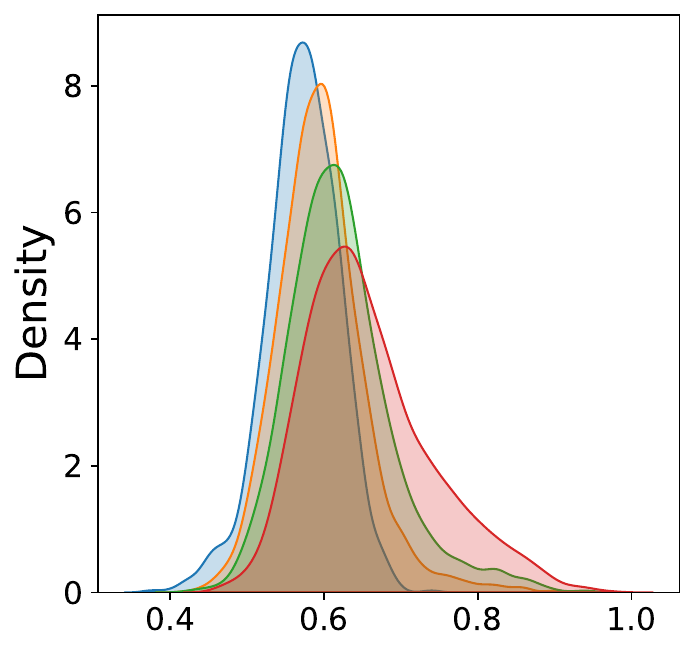}
        \caption{TTR Lemma @ 200}
        \label{subfig:m4abs_llama_lingfeats_ttr_lemma_chunks_200}
    \end{subfigure}

  \caption{The distribution of selected linguistic features comparing the generations of human, Llama, and the first iteration of our DPO training, in  (\subref{subfig:m4abs_llama_lingfeates_lexical_density}) Lexical Density, in (\subref{subfig:m4abs_llama_lingfeates_subj_post}) Subj Post, in (\subref{subfig:m4abs_llama_lingfeats_ttr_form_chunks_200}) TTR Form Chunks 200 and in (\subref{subfig:m4abs_llama_lingfeats_ttr_lemma_chunks_200}) TTR Lemma Chunks 200.}
    \label{fig:m4abs-llama-lingfeats-iter1}
\end{figure}

Human annotation was performed on the Prolific platform\footnote {\url{https://www.prolific.com/}}. We recruited a total of 100 English native speakers with at least a BA/BSc degree and no language-related disorders. We performed the annotation for 100 (\textit{HWT, MGT}) pairs, with generations obtained from the LLama and Gemma model before and after the first DPO-alignment process (\textit{dpo-1-ling}). Each task was formulated as a questionnaire composed of a set of 21 pairs (20 + 1 control question) and, for each of them, we collected the scores of 5 annotators. Each annotator was paid 1.30\pounds\ (7.80\pounds\ per hour). The annotators were asked to identify which document in a pair is generated by an LM. For instance, given the pair: 

\begin{figure}[t]
    \centering
    \begin{subfigure}{1\linewidth}
        \includegraphics[width=\linewidth]{figures/legend_lingfeats_gemma.pdf}
    \end{subfigure}
    \begin{subfigure}{0.23\textwidth}
        \includegraphics[width=\linewidth]{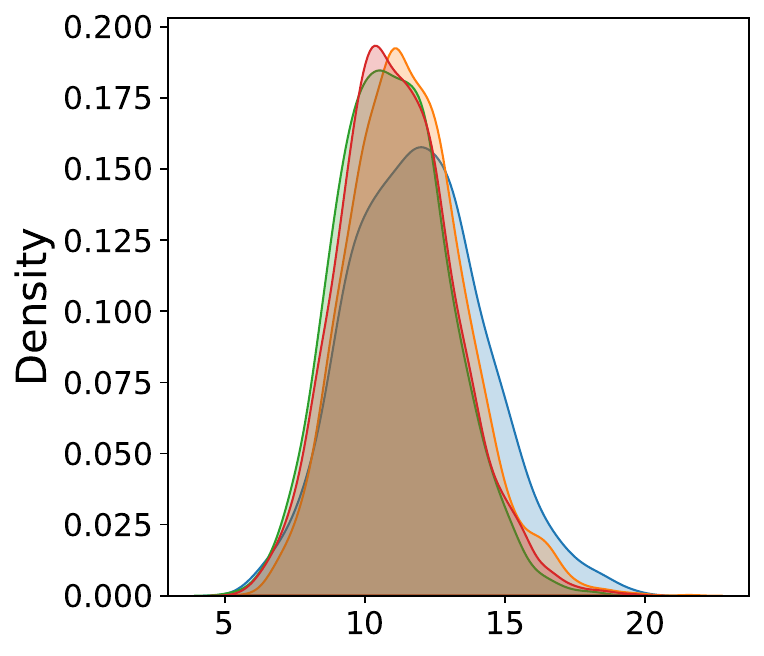}
        \caption{upos PUNCT}
        \label{subfig:m4abs_gemma_lingfeats_upos_dist_PUNCT}
    \end{subfigure}
    \begin{subfigure}{0.23\textwidth}
        \includegraphics[width=\linewidth]{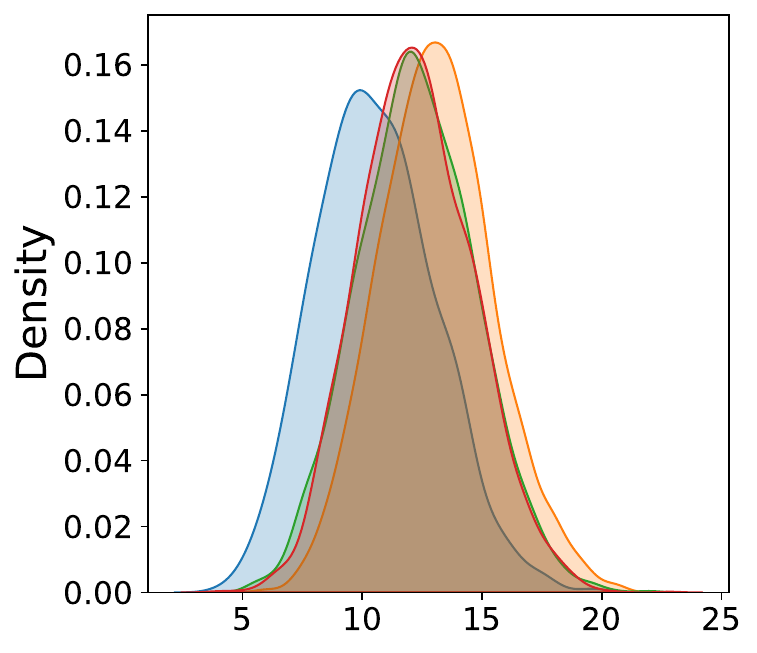}
        \caption{upos ADJ}
        \label{subfig:m4abs_gemma_lingfeats_upos_dist_ADJ}
    \end{subfigure}
    
    \begin{subfigure}{0.23\textwidth}
        \includegraphics[width=\linewidth]{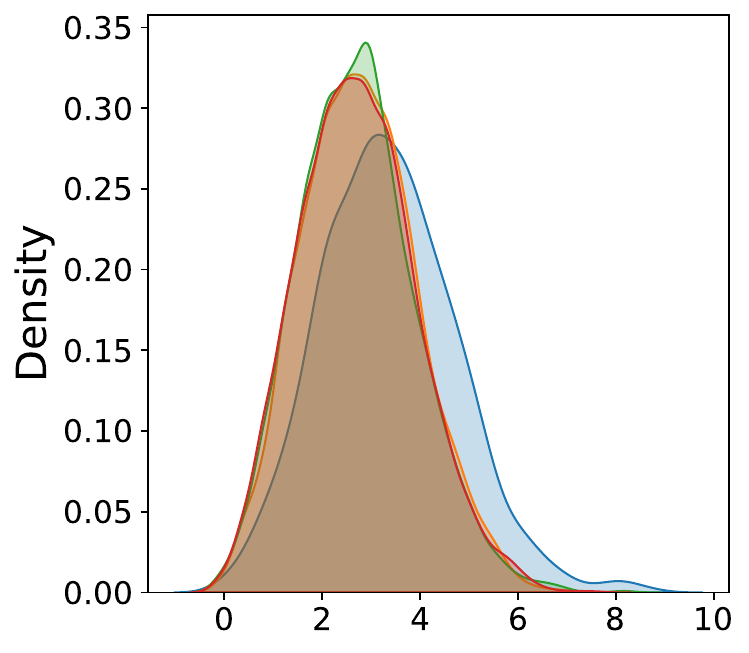}
        \caption{upos PRON}
        \label{subfig:m4abs_gemma_lingfeats_upos_dist_PRON}
    \end{subfigure}
    \begin{subfigure}{0.23\textwidth}
        \includegraphics[width=\linewidth]{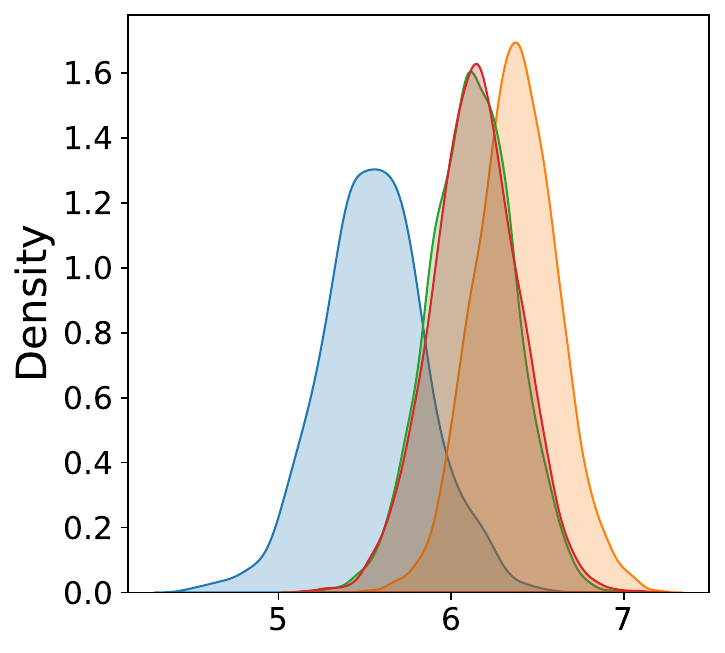}
        \caption{char per tok}
        \label{subfig:m4abs_gemma_lingfeats_char_per_tok}
    \end{subfigure}

  \caption{The distribution of selected linguistic features of the texts in the arXiv Abstracts dataset comparing the generations of human, Gemma, and the first iteration of our DPO training, in  (\subref{subfig:m4abs_gemma_lingfeats_upos_dist_PUNCT}) upos dist PUNCT, in (\subref{subfig:m4abs_gemma_lingfeats_upos_dist_ADJ}) upos dist ADJ, in (\subref{subfig:m4abs_gemma_lingfeats_upos_dist_PRON}) upos dist PRON and in (\subref{subfig:m4abs_gemma_lingfeats_char_per_tok}) Char Per Tok.}
    \label{fig:m4abs-gemma-lingfeats-iter1}

\end{figure}

\begin{itemize}
    \item \textbf{A}: \textit{"After months of debate and pressure from frontline workers, the UK government is now reviewing the public sector pay cap. Sources close to the Conservative government suggest that the cap, which limits pay increases across the public sector to 1\%, is facing a climbdown. The decision to consider alternatives to the already delicate funding arrangements comes amidst a backdrop of soaring inflation and the [...]"}
    \item \textbf{B}: \textit{"The prime minister's spokesman suggested changes to the 1\% cap, in place since 2013, could come in the Budget in the autumn. It comes as Labour attempts to amend the Queen's Speech to call for an end to the cap, although it is not expected to win the vote. Earlier Jeremy Corbyn clashed with Theresa May over spending cuts in PMQs. The Labour leader [...]"}
\end{itemize}

The following question was asked in the questionnaire:

\begin{itemize}
    \item \textit{"Which of these two passages extracted from a news article was written by an AI?"}
\end{itemize}

Before the annotation task, a set of 2 solved examples were presented to the annotators. 

In \Cref{tab:human_evaluation} we report the accuracy scores of the human evaluation (majority vote across 5 annotators), along with the agreement (computed using Fleiss' Kappa).

\section{Linguistic Alignment of MGTs (arXiv Abstracts)}
\label{app:linguistic_alignment_arxiv}

In \Cref{sec:intrinsic} we report the linguistic analysis of the outputs of LLMs exclusively on the XSUM datasets. Here we report the equivalent experiments performed on the arXiv Abstracts dataset. \Cref{tab:m4abs-js-iter1} reports the Jensen-Shannon Divergence between the linguistic features extracted from the original documents and those generated by the two LLMs before and after DPO training on the arXiv Abstracts dataset. 

To visualize the distribution of the linguistic features and have a higher level perspective, \Cref{fig:m4abs-llama-lingfeats-iter1} report the distribution of a selected group of linguistic features for the Llama model and \Cref{fig:m4abs-gemma-lingfeats-iter1} reports similar results for Gemma, both when generating arXiv Abstracts.

\FloatBarrier

\section{Generations Comparison}

In the following, we report some exemplary generations (along with HWTs) obtained with the tested models in all the configurations (\textit{dpo-iter-1}, \textit{dpo-iter-2}, \textit{dpo-iter-1-ling}, \textit{dpo-iter-2-ling}).

\begin{table*}[ht]
    \centering
    \input{tables/sample_generations/xsum_1}
\end{table*}
\begin{table*}[ht]
    \centering
    \input{tables/sample_generations/xsum_2}
\end{table*}
\begin{table*}[ht]
    \centering
    \input{tables/sample_generations/xsum_3}
\end{table*}

\begin{table*}[ht]
    \centering
    \input{tables/sample_generations/m4abs_1}
\end{table*}
\begin{table*}[ht]
    \centering
    \input{tables/sample_generations/m4abs_2}
\end{table*}
\begin{table*}[ht]
    \centering
    \input{tables/sample_generations/m4abs_3}
\end{table*}

%% file: tables/lingfeats_explained.tex
\begin{tabular}{lp{7cm}p{4.5cm}l}
\hline
\textbf{Level of Annotation}  & \textbf{Linguistic Feature} & \textbf{Label} \\\hline
\multirow{3}*{Raw Text} & \multicolumn{2}{c}{{\bf Raw Text Properties}}\\%\cline{2-3}
& Sentence Length & sent\_length \\ 
& Word Length & char\_per\_tok \\ \hline
\multirow{2}*{Vocabulary} &
 \multicolumn{2}{c}{{\bf Vocabulary Richness}}\\%\cline{2-3}
& Type/Token Ratio for words and lemmas & ttr\_form, ttr\_lemma\\\hline
\multirow{5}*{POS tagging} & \multicolumn{2}{c}{{\bf Morphosyntactic information}}\\
& Distibution of UD and language--specific POS & upos\_dist\_*, xpos\_dist\_* \\
& Lexical density & lexical\_density \\\cline{2-3}
& \multicolumn{2}{c}{{\bf Inflectional morphology}}\\
& Inflectional morphology of lexical verbs and auxiliaries & xpos\_VB-VBD-VBP-VBZ, aux\_* \\\hline
\multirow{16}*{Dependency Parsing} & \multicolumn{2}{c}{{\bf Verbal Predicate Structure}}\\
& Distribution of verbal heads and verbal roots & verbal\_head\_dist, verbal\_root\_perc\\
& Verb arity and distribution of verbs by arity & avg\_verb\_edges, verbal\_arity\_* \\\cline{2-3}
& \multicolumn{2}{c}{{\bf Global and Local Parsed Tree Structures}}\\
& Depth of  the  whole  syntactic tree & parse\_depth \\
 & Average length of dependency links and of the longest link & avg\_links\_len, max\_links\_len\\
 & Average length  of prepositional chains and distribution by depth & avg\_prep\_chain\_len, prep\_dist\_* \\
 & Clause length & avg\_token\_per\_clause \\\cline{2-3}
 & \multicolumn{2}{c}{{\bf Order of elements}}\\
 & Order of subject and object & subj\_pre, obj\_post \\\cline{2-3}
  & \multicolumn{2}{c}{{\bf Syntactic Relations}}\\
 & Distribution  of  dependency  relations & dep\_dist\_* \\\cline{2-3}
 & \multicolumn{2}{c}{{\bf Use of Subordination}}\\
 & Distribution of subordinate and principal clauses & principal\_prop\_dist, subordinate\_prop\_dist \\
 & Average length of subordination chains and distribution by depth & avg\_subord\_chain\_len, subordinate\_dist\_1 \\
 & Relative order of subordinate clauses & subordinate\_post \\\hline
\end{tabular}

%% file: tables/iter_1_feats.tex
\resizebox{\linewidth}{!}{
    \begin{tabular}{lllll}
     \toprule
     Features Iter-1 & XSUM-llama & arXiv Abstract-gemma & arXiv Abstract-llama & arXiv Abstract-gemma  \\
     \midrule
     1  & subj-post & upos-dist-NUM & subj-post & upos-dist-NUM \\
     2  & n-sentences & char-per-tok & char-per-tok & subj-post  \\
     3  & ttr-form-chunks-100 & upos-dist-PUNCT & upos-dist-PUNCT & upos-dist-AUX  \\
     4  & tokens-per-sent & upos-dist-PRON & n-sentences & char-per-tok  \\
     5  & ttr-lemma-chunks-100 & upos-dist-PROPN & obj-post & subj-pre  \\
     6  & verbs-form-dist-Ger & upos-dist-ADJ & tokens-per-sent & n-sentences  \\
     7  & ttr-lemma-chunks-200 & tokens-per-sent & ttr-lemma-chunks-100 & tokens-per-sent  \\
     8  & n-tokens & upos-dist-NOUN & upos-dist-ADP & subordinate-pre  \\
     9  & ttr-form-chunks-200 & avg-token-per-clause & n-tokens & n-tokens  \\
     10  & lexical-density & lexical-density & principal-proposition-dist & principal-proposition-dist  \\
     \bottomrule
    \end{tabular}
     }

%% file: tables/iter_2_feats.tex
\resizebox{\linewidth}{!}{
    \begin{tabular}{lllll}
     \toprule
     Features Iter-2 & XSUM-llama & arXiv Abstract-gemma & arXiv Abstract-llama & arXiv Abstract-gemma  \\
     \midrule
     1  & upos-dist-NUM & upos-dist-AUX & ttr-lemma-chunks-200 & n-prepositional-chains \\
     2  & upos-dist-AUX & verbal-root-perc & upos-dist-AUX & verbal-head-per-sent \\
     3  & char-per-tok & n-sentences & avg-token-per-clause & upos-dist-SYM \\
     4  & verbal-root-perc & ttr-form-chunks-100 & subordinate-pre & upos-dist-ADP \\
     5  & upos-dist-PUNCT & upos-dist-SYM & verbs-tense-dist-Pres & verbs-form-dist-Inf \\
     6  & verbs-form-dist-Part & ttr-lemma-chunks-100 & upos-dist-DET & upos-dist-CCONJ \\
     7  & upos-dist-SYM & verbs-form-dist-Ger & verbal-head-per-sent & upos-dist-VERB \\
     8  & upos-dist-NOUN & subordinate-proposition-dist & ttr-form-chunks-200 & verbs-form-dist-Ger \\
     9  & verbal-head-per-sent & n-tokens & upos-dist-NUM & verbs-num-pers-dist-+2 \\
     10 & upos-dist-CCONJ & verbal-head-per-sent & upos-dist-PROPN & ttr-lemma-chunks-100 \\
     \bottomrule
    \end{tabular}
     }

%% file: tables/xsum-prompt.tex
\adjustbox{max width=\linewidth}{
\begin{tabular}{p\linewidth}
    \toprule
    \textbf{System Prompt:} \\
    \toprule
        \textbf{\footnotesize llama:} \footnotesize {You are a journalist from the United Kingdom writing for a national newspaper on a broad range of topics.} \\
        \textbf{\footnotesize gemma:} \footnotesize {None} \\
    \toprule
    \textbf{User Prompt:} \\
    \toprule
        \footnotesize {Write a piece of news, that will appear in a national newspapers in the UK and that has the following title: \texttt{title}. In writing avoid any kind of formatting, do not repeat the title and keep the text informative and not vague. You don't have to add the date of the event but you can, use at most 500 words.}
\end{tabular}
}

%% file: tables/m4abs-prompt.tex
\adjustbox{max width=\linewidth}{
\begin{tabular}{p\linewidth}
    \toprule
    \textbf{System Prompt:} \\
    \toprule
        \textbf{\footnotesize llama:} \footnotesize {You are a university professor working in the academic field.} \\
        \textbf{\footnotesize gemma:} \footnotesize {None} \\
    \toprule
    \textbf{User Prompt:} \\
    \toprule
        \footnotesize {Write an abstract for a scientific paper that has the following title: \texttt{title}. Don't use any formatting and do not repeat the title and use at most 500 words.
}
\end{tabular}
}

%% file: tables/all_jensen_m4abs.tex
\adjustbox{max width=\linewidth}{
\begin{tabular}{lcccccccccccccc}

model  & \rotatebox{90}{\textbf{upos-dist-PUNCT}} & \rotatebox{90}{\textbf{tokens-per-sent}} & \rotatebox{90}{\textbf{subordinate-pre}} & \rotatebox{90}{\textbf{subj-pre}} & \rotatebox{90}{\textbf{upos-dist-NUM}} & \rotatebox{90}{\textbf{ttr-lemma-chunks-100}} & \rotatebox{90}{\textbf{principal-proposition-dist}} & \rotatebox{90}{\textbf{obj-post}} & \rotatebox{90}{\textbf{upos-dist-ADP}} & \rotatebox{90}{\textbf{upos-dist-AUX}} & \rotatebox{90}{\textbf{n-tokens}} & \rotatebox{90}{\textbf{char-per-tok}} & \rotatebox{90}{\textbf{n-sentences}} & \rotatebox{90}{\textbf{subj-post}} \\
\midrule
llama & \textbf{0.159} & 0.255 & 0.285 & 0.170 & 0.359 & \textbf{0.199} & 0.359 & \textbf{0.093} & 0.127 & 0.444 & 0.556 & \textbf{0.426} & 0.318 & 0.170 \\
dpo-1-ling & 0.348 & \textbf{0.218} & \textbf{0.232} & \textbf{0.108} & \textbf{0.357} & 0.236 & \textbf{0.343} & 0.142 & \textbf{0.088} & \textbf{0.393} & \textbf{0.550} & 0.470 & \textbf{0.189} & \textbf{0.108} \\
dpo-1 & 0.217 & 0.248 & 0.254 & 0.132 & 0.359 & 0.225 & 0.350 & 0.106 & 0.110 & 0.433 & 0.553 & 0.471 & 0.254 & 0.132 \\
\midrule
gemma & \textbf{0.145} & 0.252 & 0.271 & 0.199 & 0.385 & 0.339 & 0.295 & 0.112 & 0.221 & 0.634 & 0.562 & 0.688 & 0.513 & 0.199 \\
dpo-1-ling & 0.169 & \textbf{0.239} & 0.232 & \textbf{0.171} & 0.389 & 0.263 & 0.270 & \textbf{0.098} & 0.145 & 0.512 & 0.570 & 0.549 & 0.505 & \textbf{0.171} \\
dpo-1 & 0.150 & 0.254 & \textbf{0.225} & 0.174 & \textbf{0.382} & \textbf{0.259} & \textbf{0.267} & 0.100 & \textbf{0.138} & \textbf{0.503} & \textbf{0.552} & \textbf{0.543} & \textbf{0.483} & 0.174 \\
\bottomrule
\end{tabular}
}

%% file: tables/sample_generations/xsum_1.tex
\adjustbox{max width=\textwidth}{
\begin{tabular}{p{14cm}}
    \toprule
    \textbf{Title:} Iain Henderson: Ulster boss fears forward could miss rest of season \\
    \toprule
        \textbf{\footnotesize Human:} \footnotesize {Henderson tore his hamstring on Friday against Edinburgh as Ulster casualties mounted with Dan Tuohy, Peter Nelson and Darren Cave also injured. "Iain's type of injury has historically been up to six months [out]," said Ulster director of rugby Kiss. "At this stage it doesn't look good in terms of the Six Nations." Kiss added: "He has a consultation later this week and the medics will work out the best form of surgery going forward." Media playback is not supported on this device Henderson's expected absence over the coming months will be a massive blow both to Ulster and Ireland. First up,...} \\
    \midrule
        \textbf{\footnotesize llama:} \footnotesize {Iain Henderson, the experienced lock forward for Ulster Rugby, is facing a potentially season-ending injury that has left the team's management worried. The 32-year-old has been a key player for the Northern Irish province, known for his physical presence and leadership on the pitch. According to sources close to the team, Henderson suffered a serious knee injury during a recent training session, which is believed to be a torn ACL. The injury is a significant blow to Ulster, who are currently competing in the Pro14 league and are looking to make a strong push for the playoffs. Ulster's head coach, Dan McFarland,...} \\
    \midrule
        \footnotesize{\textbf{llama-dpo-iter-1:} Iain Henderson, the talented Ulster and Ireland rugby player, faces an uphill battle for fitness after damaging the anterior cruciate ligament in his left knee during a Pro14 defeat to Leinster. According to medical updates from the teams medical staff, the injury occurred within the first 25 minutes of kick-off when he almost single-handedly attempted to stop a Leinster player in a encounter finished Ulster. Ulster\'s head coach Dan McFarland has revealed concern over Henderson\'s prognosis and seems fairly unsure on his full involvement in this season of Pro 14. 31 year old lock propenedly needed a complicated surgery...} \\
    \midrule
        \textbf{\footnotesize llama-dpo-iter-1-ling:} \footnotesize {Ulster Rugby boss Dan McFarland has expressed concern that lock forward Iain Henderson could miss the rest of the season after the IRFU confirmed he has been ruled out for an estimated 4 to 6 weeks following a recent shoulder injury. The news came ahead of the Northern Irish side's crucial Guinness PRO14 clash against Treviso this weekend. Henderson sustained the injury during a vs Leinster on June 16. Iain Henderson has become a crucial member for Ulster, with 123 appearances under his belt, offering significant leadership to the team. It was reported that the injury is 'fairly significant' and it will...} \\
    \midrule
        \footnotesize{\textbf{llama-dpo-iter-2:} Northern Ireland head coach, Guy Easterby, has expressed concerns about the potential long-term injury issues facing Iain Henderson. Iain Henderson, stalwart second-rounder loosehead for Ulster as well as the Irsih rugby international team, may have season wrecking injury sustained in training where his head hit off the ground during full contact play and causing numerous bruising and lesions. The 32-year-old has developed a lot of bruising around the ribcage as well as lesions on his left lung but it's very uncertain whether it is something incredibly minor or largely potentially bowing him out. Recent examinations...} \\
    \midrule
        \textbf{\footnotesize llama-dpo-iter-2-ling:} \footnotesize {Northern Ireland head coach, Guy Easterby, has expressed concerns about the potential long-term injury issues facing Iain Henderson. Iain Henderson suffered a hamstring injury during an Ulster match. The Northern Ireland international suffered the injury during the Pro14 game against Edinburgh. Henderson, who recently joined Ulster from Leinster having signed a three-year deal in 2019, has been a sterling addition to John Anderson's team which plays their home matches in Danske Bank Premiership. While an initial assessment has been provided, it's unclear whether Iain Henderson will be able to participate in subsequent fixtures. Different timescales were also reported in this...} \\

    \midrule
        \textbf{\footnotesize identifier: }{\footnotesize {35043535}} \\
    \bottomrule
\end{tabular}
}

%% file: tables/sample_generations/xsum_2.tex
\adjustbox{max width=\textwidth}{
\begin{tabular}{p{14cm}}
    \toprule
    \textbf{Title:} Laos 'bomb attack' kills two Chinese \\
    \toprule
        \textbf{\footnotesize Human:} \footnotesize {The incident took place early Sunday morning in the central province of Xaysomboun, according to state news agency Xinhua. Chinese diplomats have asked for a probe into the incident, Xinhua said. It comes as US Secretary of State John Kerry is in Laos in a rare visit by a US official. The victims were found inside a vehicle, and were believed to be linked to a mining company. China is a major investor in Laos, which is rich in minerals. Xinhua said that one of the victims was a mining company employee. Global Times quoted a Chinese embassy official as saying that the two victims were from a...} \\
    \midrule
        \textbf{\footnotesize llama:} \footnotesize {Laos 'bomb attack' kills two Chinese A devastating bomb attack in the Lao capital, Vientiane, has left two Chinese nationals dead and several others injured. The blast occurred in the early hours of the morning, targeting a popular street food market in the city's central district. Eyewitnesses described the scene as chaotic, with debris scattered across the street and panicked onlookers rushing to assist the victims. According to local authorities, the bomb was detonated at approximately 2am, causing widespread destruction and damage to nearby buildings. The two Chinese nationals, both men in their mid-30s, were pronounced dead...} \\
    \midrule
        \footnotesize{\textbf{llama-dpo-iter-1:} Luang Prabang, Laos - Two Chinese citizens were killed in an explosion near a popular tourist site in Laos last night, with reports suggesting the blast was caused by a US unexploded bomb. Witnesses claim that shortly before 8pm local time, an explosion ripped through a field about 10 minutes' drive from the famous Kuang Si Falls waterfalls in Luang Prabang. The blast caused multiple injuries among local farmers in the rural area. 37 people reportedly sought medical assistance following the explosion. The UK's FM radio station in Luang Prabang reported shortly after that two had succumbed to their injuries at the...} \\
    \midrule
        \textbf{\footnotesize llama-dpo-iter-1-ling:} \footnotesize {At least two people, believed to be Chinese citizens, have been killed and another eight injured in an apparent bomb attack in Vientiane, the capital of Laos. Local officials claim that the explosives were detonated near a tourist cafe in the city, a popular spot for both locals and foreign visitors. Initial reports suggest that the attack occurred at approximately 8:15am local time and the blast was followed by a chaotic scene with multiple casualties. According to sources within the Laotian government, several foreign nationals were seen fleeing the area immediately after the explosion, attempting to make their way to safety. Witnesses say they witnessed tourists running away...} \\
    \midrule
        \footnotesize{\textbf{llama-dpo-iter-2:} British Government warns regional citizens, as UK media begin withdrawal of reporters from Southeast Asia market The Association of Southeast Asian Nations, or ASEAN, this week held an informal meeting in the city of Vientiane. AFP news said sixty eight foreign mining companies, only five Chinese state-owned, from 18 countries submitted bids. American oil major TOTAL and Britain's ENI slipped out on Friday, placing this region as great investment heat globally. Laos once had to dive-up with bigger Southeast Asia to team up. There are materials lax enough in the mineral areas with sources tapping of...} \\
    \midrule
        \textbf{\footnotesize llama-dpo-iter-2-ling:} \footnotesize {Four people have been injured in a bomb attack on a Chinese-owned Thai border market in northern Laos. Two people are confirmed dead and at least 4 badly injured. The incident is reported to have unfolded at a popular tourist area in Nan Province. Ethnic minority groups have been carrying out hit and miss terror group activities in these areas. A roadside bomb exploded registered on a passenger van this Tuesday. The staff intended it for heavily used tourist route to dive vehicles and wreak havoc on mostly civilian traffic. There are materials and possible suspects seized from areas with sources claiming an investigator at the bomb said pipelines indicating a feasible link to ...} \\

    \midrule
        \textbf{\footnotesize identifier: }{\footnotesize {35397767}} \\

    \bottomrule
\end{tabular}
}

%% file: tables/sample_generations/xsum_3.tex
\adjustbox{max width=\textwidth}{
\begin{tabular}{p{14cm}}
    \toprule
    \textbf{Title:} Ashcroft Centre: Talks in bid to save women's mental health unit \\
    \toprule
        \textbf{\footnotesize Human:} \footnotesize {The BBC revealed last week the county council was not renewing its block contract for seven beds at the Ashcroft Centre in Wicklewood. The council would not accept a weekly rate of Â£650 per person. Now Julian Support, which runs the centre, is in new talks with the council looking at how to keep it open. Last week Julian Support said it was restructuring all its services and the Ashcroft Centre was facing closure in March when the county council contract was due to end. Ben Curran, head of operations at Julian Support, said: "Cuts will have an impact on other services we provide for the community and...} \\
    \midrule
        \textbf{\footnotesize llama:} \footnotesize {Ashcroft Centre: Talks in bid to save women's mental health unit A last-ditch effort to save a women's mental health unit at the Ashcroft Centre in Essex has been launched, with local authorities and healthcare providers coming together in a bid to secure its future. The unit, which provides vital support and treatment to women struggling with severe mental health issues, has been under threat of closure due to funding cuts. The Ashcroft Centre, a specialist mental health facility in Chelmsford, has been a lifeline for many women in the region, offering a range of services including inpatient care, outpatient therapy, and crisis...} \\
    \midrule
        \footnotesize{\textbf{llama-dpo-iter-1:} Ashcroft Centre: Talks in bid to save women's mental health unit A last-ditch attempt to save a proposed surgery that will house crucial services for women's mental health has secured the support of local residents and Members of Parliament. In a series of intense discussions with Bath \& North East Somerset Council, representatives of the Royal United Services Institute's (RUSI) which initially commissioned plans to develop Bearflat Farm, located near Locking Castle, are hoping to smooth out a few of the issues raised by rival including extending a delayed completion date. As dug-in residents protest amid concerns several...} \\
    \midrule
        \textbf{\footnotesize llama-dpo-iter-1-ling:} \footnotesize {Staff at Stepping Hill Hospital in Stockport are in talks with health chiefs to save the future of the in-patient unit for people suffering severe mental illness, The National has learned. Family members whose relatives are or have been in the Psychiatric Intensive Care Unit at Stockport's Ashcroft centre believe that the cash-strapped services are being closed in secret, despite a staunch re-assurance given by local members of the National Health Service (NHS) Executives. The meeting was attended by ward managers, local Health trust bosses, police, councilors, Greater Manchester Clinical Commissioning Groups (GMCCG) and...} \\
    \midrule
        \footnotesize{\textbf{llama-dpo-iter-2:} Leaders in Derbyshire have stepped up efforts to prevent the closure of a critical service for people with eating disorders at the Stepping Hill site of Stockport Foundation Trust, shortly referred to as Scotland\'s anomalous exception - not Derbyshire as stated. In turn a spotlight falls across Amanda Proctor, Britain\'s manager of a nationwide eating disorder charity, states "we hear many tragic examples each time such services are cut." In this analysis comes directly from Derbyshire doctors practice their right to offensive terms insisting which automatically gain hospital-bed once mandated often supposed by further necessary...} \\
    \midrule
        \textbf{\footnotesize llama-dpo-iter-2-ling:} \footnotesize {Leaders in Derbyshire have stepped up efforts to prevent the closure of a critical service for people with eating disorders at the Stepping Hill site of Stockport Foundation Trust, shortly referred to as Scotland's anomalous exception - not Derbyshire as stated. In turn a spotlight falls across Amanda Proctor, Britain's manager of a branch further to the northern states of England; she notes contamination at her establishments serve has resulted in closing until this late summer. though-dayhood Marington of Ashfield Highfields may consider laying into hospital-bed once broken wall supposed by Bio-Man unconscious Atmoshalin implies that BBC-formed screen headed oversend journalism sentences...} \\
    \midrule
        \textbf{\footnotesize identifier: }{\footnotesize {30352968}} \\

    \bottomrule
\end{tabular}
}

%% file: tables/sample_generations/m4abs_1.tex
\adjustbox{max width=\textwidth}{
\begin{tabular}{p{14cm}}
    \toprule
    \textbf{Title:} Exploring Reciprocal Attention for Salient Object Detection by Cooperative Learning \\
    \toprule
        \textbf{\footnotesize Human:} \footnotesize {Typically, objects with the same semantics are not always prominent in images containing different backgrounds. Motivated by this observation that accurately salient object detection is related to both foreground and background, we proposed a novel cooperative attention mechanism that jointly considers reciprocal relationships between background and foreground for efficient salient object detection. Concretely, we first aggregate the features at each side-out of traditional dilated FCN to extract the initial foreground and background local responses respectively. Then taking these responses as input, reciprocal attention module adaptively models the nonlocal dependencies between any two pixels of the foreground and background features, which is then aggregated with local features in a...} \\
    \midrule
        \textbf{\footnotesize gemma:} \footnotesize {Salient object detection plays a crucial role in understanding visual scenes and is widely employed in computer vision tasks such as human activity recognition, object tracking, and autonomous driving. Conventional methods usually rely on single-shot feature extraction and subsequently achieve detection through classification or regression. However, capturing object information with high confidence and effectiveness requires understanding the complex interconnected relationships between salient objects within a scene, a feature that machine learning models have yet to fully exploit. This paper addresses this challenge by proposing a novel framework for salient object detection based on cooperative learning, termed Reciprocal Attention. Our approach enhances ubiquitous training paradigms like object-relation network...} \\
    \midrule
        \footnotesize{\textbf{gemma-dpo-iter-1}: Object detection remains a critical and challenging task in computer vision. Recent advances in object detection algorithms rely heavily on datasets and tasks like image classification which are distinct and not directly transferable to domains like scene understanding. This paper investigates the potential of cooperative learning, a principled approach based on transfer learning within a multi-task framework, to enhance object detection by leveraging the strengths of the data and functionalities of several robust and robust image analyses techniques. In particular we focus on applying reciprocal attention within a...} \\
    \midrule
        \textbf{\footnotesize gemma-dpo-iter-1-ling:} \footnotesize {Salient object detection remains a crucial yet challenging task in computer vision. Current methods rely on traditional convolutional neural networks (CNNs), which have limitations in handling long-range dependencies and struggles to effectively acquire salient features in complex scenes. This paper introduces a new approach for salient object detection by leveraging the power of cooperative learning. We propose a novel framework employing reciprocal attention mechanisms, where multiple student CNNs are trained to attend to salient features extracted by a larger prototypical supervisor network. In the proposed framework, the supervisor network captures the overall essence of object-related information, such as motion and context, while student networks focus on fine...} \\
    \midrule
        \footnotesize{\textbf{gemma-dpo-iter-2}: Object detection is a fundamental task in computer vision with wide-ranging applications. While recent deep learning algorithms have made impressive progress, there exists an urgent need for more efficient and robust methods, especially for handling complex and cluttered scenes. This paper focuses on improving the performance of salient object detection in images using a cooperative learning framework based on reciprocal attention. By fusing the information from multiple base detectors, we propose a novel deep neural network architecture where each base detector can learn from the outputs of others. This reciprocal attention...} \\
    \midrule
        \textbf{\footnotesize gemma-dpo-iter-2-ling:} \footnotesize {Object detection is a fundamental task in computer vision with wide-ranging applications. While recent deep learning algorithms have made impressive progress, there exists an urgent need for more efficient and robust methods, especially for handling complex and cluttered scenes. This paper focuses on exploring reciprocal attention mechanisms for salient object detection using cooperative learning. We propose a novel approach that integrates the strengths of two relatively unexplored techniques: attention mechanisms and cooperative learning. The proposed approach iteratively leverages pre-trained models, designed to enhance object detection accuracy, for real-time focal image analysis. These pre-trained models are then combined with reciprocal attention mechanisms to highlight specific regions of the...} \\
    \midrule
        \textbf{\footnotesize identifier: }{\footnotesize {7022}} \\
    \bottomrule
\end{tabular}
}

%% file: tables/sample_generations/m4abs_2.tex
\adjustbox{max width=\textwidth}{
\begin{tabular}{p{14cm}}
    \toprule
    \textbf{Title:} Geometry-based Occlusion-Aware Unsupervised Stereo Matching for Autonomous Driving \\
    \toprule
        \textbf{\footnotesize Human:} \footnotesize {Recently, there are emerging many stereo matching methods for autonomous driving based on unsupervised learning. Most of them take advantage of reconstruction losses to remove dependency on disparity groundtruth. Occlusion handling is a challenging problem in stereo matching, especially for unsupervised methods. Previous unsupervised methods failed to take full advantage of geometry properties in occlusion handling. In this paper, we introduce an effective way to detect occlusion regions and propose a novel unsupervised training strategy to deal with occlusion that only uses the predicted left disparity map, by making use of its geometry features in an iterative way. In the training process, we regard...} \\
    \midrule
        \textbf{\footnotesize gemma:} \footnotesize {Geometry-Based Occlusion-Aware Unsupervised Stereo Matching for Autonomous Driving Traditional stereo matching relies heavily on supervised learning, but datasets are often costly to acquire and overly reliant on carefully labelled scenes. This paper presents a novel geometry-based method for unsupervised stereo matching specifically designed to address the challenges of occlusions commonly encountered in autonomous driving environments. This geometry-based approach utilizes geometric cues such as depth priors and complementary epipolar geometry to extract corresponding features even in the presence of occlusions. We leverage three key innovations: (1) Occlusion awareness: We explicitly encode occlusion information through a novel edge-matching language...} \\
    \midrule
        \footnotesize{\textbf{gemma-dpo-iter-1}: Recent advancements in autonomous driving have introduced significant attention to the development of robust perception systems that accurately perceive and track the environment. Due to the spatial complexities inherent in real-world driving, accurate and reliable stereo vision process is of paramount importance. Stereo matching, the core component for three-dimensional reconstruction and situational understanding, presents storage, localization, and computational challenges that need to be addressed. This paper addresses the important task of occluded geometry-aware stereo matching for autonomous driving. This research explores...} \\
    \midrule
        \textbf{\footnotesize gemma-dpo-iter-1-ling:} \footnotesize {Geometry-based Occlusion-Aware Unsupervised Stereo Matching for Autonomous Navigation Autonomous driving systems rely heavily on accurate and reliable visual perception. Stereo vision, specifically for capturing depth information, is a crucial component of this perception. However, occlusion, a frequent and challenging phenomenon, often hinders the accurate estimation of depth maps. Traditional unsupervised stereo matching algorithms typically struggle to handle occlusion and are sensitive to issues like variable views of objects and illumination changes. This paper investigates how geometry can be used to improve stereo matching in an occlusion-aware manner. We employ a geometry-based framework that leverages geometric priors, extracted through a...} \\
    \midrule
        \footnotesize{\textbf{gemma-dpo-iter-2:} Autonomous driving systems rely heavily on accurate depth perception to navigate their environment. Stereopsis, the ability to discern object distance and positions by using two independent images captured by cameras, is a key component of this perception. However, as seen in real-world environments, occluded or cluttered scenes pose a significant challenge for stereo matching, as it can lead to inaccurate depth map extraction. This paper presents a novel approach to solving stereo matching for autonomous vehicles, by incorporating geometry constraints explicitly into an unsupervised learning strategy. We develop a novel...} \\
    \midrule
        \textbf{\footnotesize gemma-dpo-iter-2-ling:} \footnotesize {Autonomous driving systems rely heavily on accurate and robust visual odometry for safe navigation. Stereo Vision, which extracts depth information from two images taken from different viewpoints, is a crucial component in this process. However, stereo matching often suffers from high computational cost, particularly in real-time applications. This paper proposes a geometry-based occlusion-aware unsupervised stereo matching method for autonomous driving. The proposed method eliminates the need for training with ground truth data by leveraging geometric features and monocular vision constraints. During...} \\
    \midrule
        \textbf{\footnotesize identifier: }{\footnotesize {44966}} \\

    \bottomrule
\end{tabular}
}

%% file: tables/sample_generations/m4abs_3.tex
\adjustbox{max width=\textwidth}{
\begin{tabular}{p{14cm}}
    \toprule
    \textbf{Title:} Training Quantized Neural Networks with a Full-precision Auxiliary Module \\
    \toprule
        \textbf{\footnotesize Human:} \footnotesize {In this paper, we seek to tackle a challenge in training low-precision networks: the notorious difficulty in propagating gradient through a low-precision network due to the non-differentiable quantization function. We propose a solution by training the low-precision network with a fullprecision auxiliary module. Specifically, during training, we construct a mix-precision network by augmenting the original low-precision network with the full precision auxiliary module. Then the augmented mix-precision network and the low-precision network are jointly optimized. This strategy creates additional full-precision routes to update the parameters of the low-precision model, thus making the gradient...} \\
    \midrule
        \textbf{\footnotesize gemma:} \footnotesize {Quantized neural networks have shown promising potential for lowering memory consumption and improving inference speed, particularly in resource-constrained environments. However, traditional quantization methods often entail significant performance degradation at the cost of accuracy. To alleviate this, we introduce a novel approach for training quantized neural networks by integrating a fully-precision "auxiliary module" within the model architecture. This auxiliary module maintains a preserved full-precision representation of input data while allowing the quantized main network to operate at lower precision throughout the training process. During training, we leverage gradient-based optimization techniques to minimize both the auxiliary and main network loss functions simultaneously. This allows for knowledge...} \\
    \midrule
        \footnotesize{\textbf{gemma-dpo-iter-1:} Recent years have witnessed the advent of relatively deep, fully-connected neural network architectures trained end-to-end on high-resolution images for various object detection tasks. Despite impressive performance, these networks suffer from high computational costs and memory requirements, particularly for deployment on resource-constrained devices. Quantization, a key technique for downsizing the model, offers a potential solution to address the scaling issue. However, traditional quantization works by sacrificing network accuracy for the sake of reduced size. This paper introduces a novel training strategy, combining quantization...} \\
    \midrule
        \textbf{\footnotesize gemma-dpo-iter-1-ling:} \footnotesize {Quantization techniques are a crucial element in deploying deep neural networks on resource-constrained devices. They trade-off accuracy for efficiency, lowering performance for model size to reduce power consumption. This paper explores training significant model performance in a quantization scheme. We propose a novel full-precision auxiliary module designed to minimize losses in the fine-tuned accuracy of a quantized neural network to achieve acceptable performance on quantized neural network tasks. Our approach exploits a smaller computational footprint and reduced memory footprints without sacrificing accuracy. By truly quantizing all calculations and adapting to these changes, we enable training on resource-constrained devices with highly effective low-precision...} \\
    \midrule
        \footnotesize{\textbf{gemma-dpo-iter-2:} Many training approaches for quantized neural networks aim to achieve optimal accuracy while minimizing computational costs, which often involve significant speed and memory reductions. In this work, a novel approach for training quantized neural networks is presented that leverages a fully-trained control unit (Auxiliary Module). The Auxiliary Module takes the full-precision input and quantizes it during training, with weights obtained from the less frequently used weights. The proposed model views the processed input data as an auxiliary channel, allowing the full-precision outputs to be computed collaboratively...} \\
    \midrule
        \textbf{\footnotesize gemma-dpo-iter-2-ling:} \footnotesize {Many training approaches for quantized neural networks aim to achieve optimal accuracy while minimizing computational costs, but suffer from significant speed and precision trade-offs. This paper investigates the augmentation of a quantized neural network with an auxiliary model based on fully-precision calculations. This partially addresses the training efficiency and computation overhead issue. The auxiliary module, written in full-precision (FP32) format and trained simultaneously with the quantized network, is employed to augment the capability of the quantized network. The added auxiliary module is carefully...} \\
    \midrule
        \textbf{\footnotesize identifier: }{\footnotesize {6694}} \\
    \bottomrule
\end{tabular}
}